%% file: 3DVMain.tex
\documentclass[10pt,twocolumn,letterpaper]{article}

\usepackage{3dv}
\usepackage{times}
\usepackage{epsfig}
\usepackage{graphicx}
\usepackage{color}
\usepackage{subcaption}
\usepackage{comment}
\usepackage{physics}
\usepackage{float} 
\usepackage{array}
\usepackage{dsfont}
\usepackage{url}
\usepackage{textcomp} 

\usepackage{cuted}

\usepackage{enumitem}

\usepackage{amsmath,amssymb,amsthm} 

\DeclareMathOperator*{\argmin}{arg\,min}

\newtheorem{proposition}{Proposition}[section]
\newtheorem{lemma}{Lemma}[section]

\definecolor{darkorange}{rgb}{1.0, 0.55, 0.0}

\usepackage[pagebackref=true,breaklinks=true,letterpaper=true,colorlinks,bookmarks=false]{hyperref}

\usepackage[labelfont=bf]{caption} 

\threedvfinalcopy

\begin{document}

\title{Adiabatic Quantum Graph Matching with Permutation Matrix Constraints}

\author{Marcel Seelbach Benkner$^1$ \hspace{1.8em} Vladislav Golyanik$^2$ \hspace{1.8em} Christian Theobalt$^2$ \hspace{1.8em} Michael Moeller$^1$\vspace{10pt}\\
$^1$University of Siegen 
\textbf{\hspace{3em}} 
$^2$Max Planck Institute for Informatics, SIC
} 

\maketitle
\begin{abstract}
Matching problems on 3D shapes and images are challenging as they are frequently formulated as combinatorial quadratic assignment problems (QAPs) with permutation matrix constraints, which are $\mathcal{NP}$-hard. 
In this work, we address such problems with emerging quantum computing technology and 
propose several reformulations of QAPs as unconstrained problems  suitable for efficient execution on quantum hardware. 
We investigate several ways to inject permutation matrix constraints in a quadratic unconstrained binary optimization problem which can be mapped to quantum hardware. 
We focus on obtaining a sufficient \textit{spectral gap}, which further increases the probability to measure optimal solutions and valid permutation matrices in a single run. 
We perform our experiments on the quantum computer D-Wave 2000Q ($2^{11}$ qubits, adiabatic). 
Despite the observed discrepancy between simulated adiabatic quantum computing and execution on real quantum hardware, our reformulation of permutation matrix constraints increases the robustness of the numerical computations over other penalty approaches in our experiments. 
The proposed algorithm has the potential to scale to higher dimensions on  future quantum computing architectures, which opens up multiple new  directions for solving matching problems in 3D computer vision and  graphics\footnote{visit  \url{http://gvv.mpi-inf.mpg.de/projects/QGM/} (the source code is available)}. 
\end{abstract} 

\section{Introduction}

The question how quantum computing can be beneficial in solving problems in computer vision is still relatively unexplored.
A quantum computer is a computing machine which takes advantage of quantum effects, \textit{i.e.,} quantum superposition, entanglement, tunnelling and contextuality \cite{nielsen,howard2014contextuality}. 
Since late 1980s, the quantum computing paradigm attracts more and more attention of computer scientists. 
Multiple quantum methods were subsequently shown to improve the computational complexity  \cite{BernsteinVazirani1993,Simon1994,Shor,Grover96afast,vanDamSeroussi2002} compared to their classical algorithmic counterparts. 
Only recently the practicability of all these methods has been confirmed \cite{quantumSupremacy}. 
Thus, it is only a few years since quantum computing technology became mature enough and the first functional quantum computers suitable for real-world problems became available for research purposes for a broader community \cite{IBMQExp,DWAVEQPU2020,Neukart2017,coles2018quantum,pointCorrespondence}. 

\begin{figure}
    \centering
    \includegraphics[width = 0.47\textwidth]{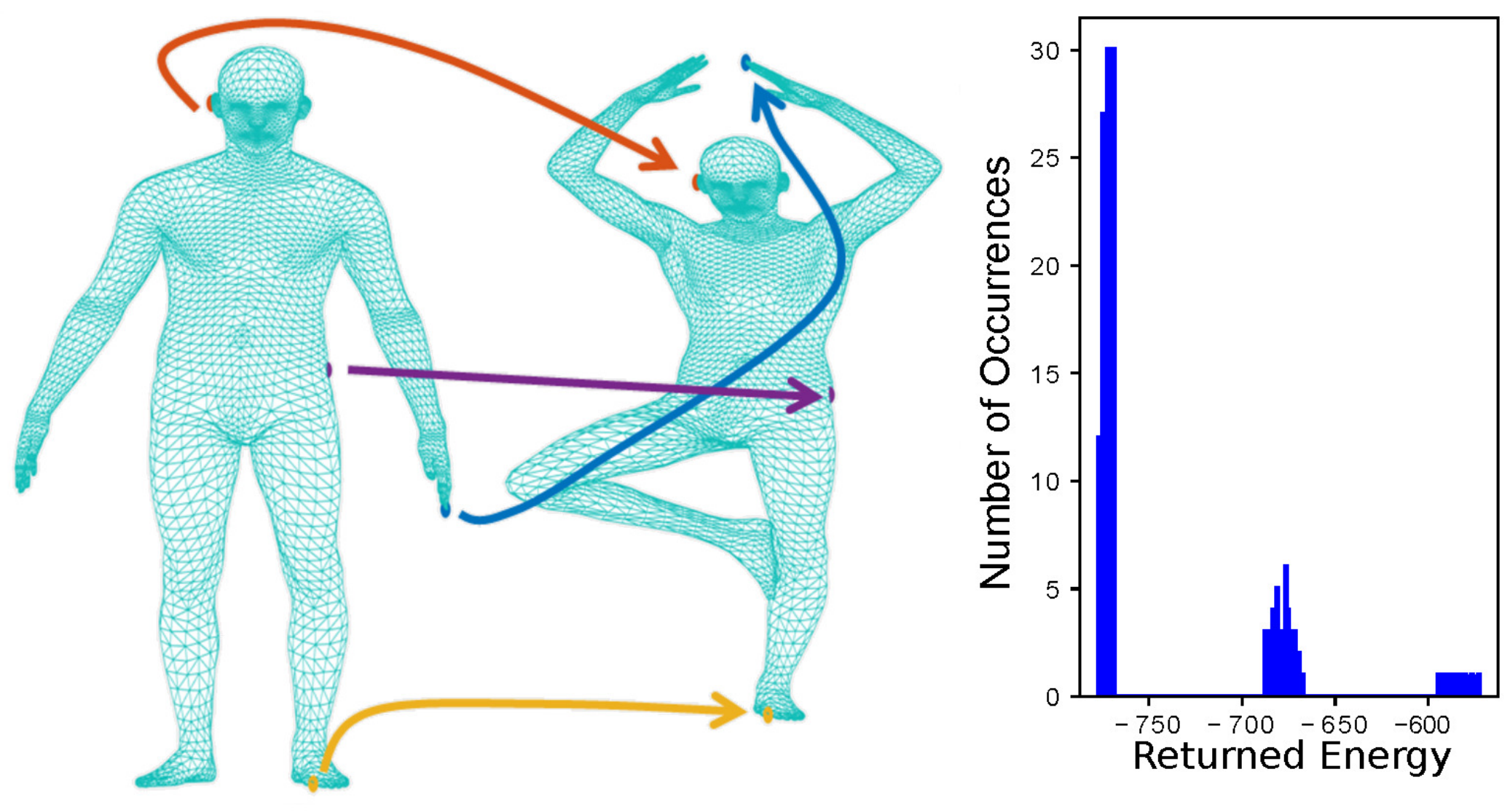}
    \caption{Example of 3D shape matching on quantum annealer D-Wave 2000Q. \textbf{(Left:)} 3D Shapes from the FAUST dataset \cite{Bogo2014} and the matches extracted with our QGM. 
    \textbf{(Right:)} Histogram of solutions on D-Wave 2000Q. 
    The peaks in the solution distribution arise from the proposed regularization which enhances the chance to measure a valid permutation matrix. 
    The left (highest) bin corresponds to valid permutations. 
    The solutions with the lowest energy, \textit{i.e.,} the left-most measured values in the histogram,  
    correspond to the recovered permutation in the picture.} 
    \label{fig:shapeMatching}
\end{figure}

Our motivation in this paper is to investigate the applicability of modern and upcoming adiabatic quantum computers (AQC-ers) for computer vision and graphics tasks. 
Therefore, we choose the fundamental problem of visual computing \textit{i.e.,} combinatorial graph matching (CGM). 
This problem allows for a general formulation and, on the other hand, is challenging enough to map on a quantum computer, for the reasons to become clear in the following. 
Many matching problems on 2D, 3D and higher-dimensional structures 
such as point set and mesh alignment \cite{Kezurer2015, Maron2016} 
require optimizing a quadratic cost function over the set of  permutation matrices, \textit{i.e.,} solving a quadratic  assignment problem (QAP) 
\begin{equation} 
\min_{X\in \mathbb{P}_n} \quad f(\mathbf{x} ):= \mathbf{x} ^{\text{T}}W \mathbf{x}  + \mathbf{c}^{T}\mathbf{x}  , \label{DASProblemIntro}
\end{equation}
where $\mathbf{x} = \text{vec}(X)$ is the $n^2$-dimensional vector corresponding to the matrix $X\in \mathbb{R}^{n\times n}$, which is constrained to lie in the set of permutation matrices $\mathbb{P}_n$. 
While there exist several well-working heuristics and approximate solvers for \eqref{DASProblemIntro} (see Sec.~\ref{sec:related} for a summary), there are problem instances 
on which every such approach would fail to provide optimal results within reasonable computational time as to be expected due to the $\mathcal{NP}$-hard nature of QAPs. 

\noindent\textbf{Motivation and Contributions.} 
With adiabatic quantum computing (AQC-ing) moving from theoretical considerations to actual implementations \cite{economist,leapoutlab}, it becomes increasingly attractive for tackling $\mathcal{NP}$-hard problems in computer vision. 
Constrained problems like \eqref{DASProblemIntro} can unfortunately not be solved with AQC-ers directly, and have to be converted to a quadratic \textit{unconstrained} binary optimization problem (QUBO). 
We observe that a straightforward reformulation of the constrained problem as a QUBO via a quadratic penalty increases the difficulty of the resulting problem significantly. 
We conjecture that this difficulty is largely related to the problem's property called the \textit{spectral gap},  \textit{i.e.,} the difference between the lowest and second-lowest energy state, and tailor our approaches at maximizing the latter while still provably solving the original constrained problem. 
To summarize, our \textbf{contributions} 
are: 
\vspace{-1pt}
\begin{itemize}[leftmargin=*] 
    \setlength{\itemsep}{0pt}
\item We show how quadratic assignment problems \eqref{DASProblemIntro} with permutation matrix constraints can be mapped to QUBOs and efficiently solved with quantum annealing for small problem instances (Sec.~\ref{sec:permutationMatrices}), see Fig.~\ref{fig:shapeMatching} for an example of 3D shape registration. 
We call our method \textit{Quantum Graph Matching} (QGM).  
It opens up new opportunities for multiple problem types in 3D computer vision, with a potentially high impact for the future generation of quantum annealers. 
We impose the permutation matrix constraints not in a post-processing step (\textit{e.g.,} by projecting a relaxed solution to the space of permutation matrices), but directly through problem weights between the qubits, which leads to high probabilities of measuring solutions corresponding to valid permutation matrices. 
\item The probability to obtain globally-optimal solutions depends on 1) how the problem is mapped to a QUBO and 2) what is the spectral gap of the mapping. We thus propose different approaches to map our formulation to a QUBO and perform spectral gap analysis (Sec.~\ref{sec:numericalResults}). 
Our new way to impose soft permutation matrix constraints is controllable by a parameter $\lambda$ per each row of the permutation matrix.
\item Numerical verification in simulated experiments as well as on a real AQC-er, which shows  
that the proposed methods effectively increase the success rate of solving combinatorial optimization problems with permutation matrix constraints (Sec.~\ref{sec:numericalResults}). 
Despite several works related to machine learning and image classification \cite{Neven2012,Adachi2015arXiv,OMalley2018,Adachi2015arXiv}, where experiments on quantum hardware were performed, and some theoretical works in quantum computing  for computer vision \cite{Neven2008arXiv,nguyen2018image,Wiebe2012,Chin2020quantum,Khoshaman2018}, 
we are not aware of previous AQC-ing experiments for image and shape matching. 
\end{itemize}

In the rest of the paper, Sec.~\ref{sec:quantumComputing} discusses the foundations of modern adiabatic quantum computing. 
Sec.~\ref{sec:related} reviews related works. 
Three variants of our QGM based on minimization of quadratic objectives over permutation matrices with AQC-ing are introduced in Sec.~\ref{sec:QGM}. 
Sec.~\ref{sec:numericalResults} elaborates on numerical results with simulated and real data, followed by a  Discussion~\ref{sec:discussion} and Conclusion~\ref{sec:conclusions}. 

\input{quantum_computing_new}
\section{Related Work}\label{sec:related} 

Our approach relates to multiple method classes proposed in the literature so far. 
In this section, we review the most related categories, \textit{i.e., } 
computer vision methods using quantum hardware including image matching and encoding of permutation matrices, as well as classical solutions to graph matching problems. 
\noindent\textbf{Quantum Computing in Computer Vision.} 
Promising potential applications of quantum computers range from data fitting  \cite{Wiebe2012,Chin2020quantum} and image recognition  \cite{Neven2008arXiv,Neven2012,OMalley2018} to training artificial neural networks  \cite{Adachi2015arXiv,Khoshaman2018}. 
Classification-related problems were addressed 
in \cite{treeCover}, to enhance the detection of vegetation zones in aerial images on D-Wave, in  \cite{OMalley2018} 
to learn facial features and reproduce facial image collections, and in  \cite{nguyen2020regression} which proposes a dictionary learning method for image  classification. 
\cite{yanquantum} covers various low-level quantum image processing topics such as quantum image encryption and segmentation. 
However, it does not address integration of permutation matrix constraints. 
To the best of our knowledge, \cite{pointCorrespondence} is the first work introducing quantum computing for computer vision at a major vision conference. 
The method finds a rotation that aligns two point sets and discretizes the space of affine transformation matrices to formulate the problem as \eqref{eq:OptProb}. 

\noindent\textbf{Quantum Computing for Matching Problems.} 
\cite{gate-based} finds a small image embedded in a larger one by adapting the Grover's algorithm \cite{grover1997quantum}. 
Neven \textit{et al.}~\cite{Neven2008arXiv} propose to match two images with AQC-ing. 
Their mapping requires finding the maximum independent vertex set of a conflict graph without a focus on permutation constraints. 
Examples of solving quadratic assignment problems with AQC-ing are \cite{stollenwerk2019flight,chooinvestigating}. 
Stollenwerk \textit{et al.}~\cite{stollenwerk2019flight} solve the flight gate assignment problem by using qubits to indicate whether resources (\textit{e.g.,} flights) are assigned to facilities (\textit{e.g.,} gates). 
The resulting matrix is indexed over the resources and facilities, and the quadratic constraints are converted to weights favoring valid configurations in the solution space. 
Choo \cite{chooinvestigating} investigates several ways to map quadratic assignment problems to 
QUBO. Their analysis leads mainly to a quadratic penalty terms. 
They also propose a decomposition approach to mitigate the challenges. 
\cite{gaitan2014graph,zick2015experimental} addresses graph isomorphism problem with AQC-ing. 
In \cite{gaitan2014graph}, permutation matrix constraints are enforced with a penalty term in the Hamiltonian. 
They first write the permutation in a table and then represent the entries in the binary system. 
Zick \textit{et al.}~\cite{zick2015experimental} 
allocate a QUBO variable for every possible pair of vertices of the same degree and solve the problem with AQC-ing, but do not use a binary representation of the permutation table.  
In this paper, we investigate a different new way of injecting permutation matrix constraints. 

\noindent\textbf{Classical Related Methods.} 
In terms of classical approaches to solve matching problems, it is well-known that problems of the form \eqref{DASProblemIntro} can be solved exactly if the costs are linear ($W=0$) using the Hungarian method \cite{Kuhn55thehungarian} or the auction algorithm \cite{Bertsekas1979}. 
For truly quadratic costs, relaxations based on message passing \cite{Kolmogorov2015} 
and Lagrange duality \cite{Swoboda_2017_CVPR,Torresani2013},  
as well as general convex approximations  \cite{SchellewaldSchnoerr2005,Aflalo2015,relaxations,Dym2017,Fogel2013,BurghardKlein2017} 
have been proposed. 
Path following \cite{Zaslavskiy2009,ZhouDelaTorre2016,Jiang2017} 
and believe propagation \cite{Anguelov2004} 
are popular heuristics to tackle \eqref{DASProblemIntro} directly, and we refer to \cite{Pardalos94} 
for an overview of combinatorial approaches to the QAP. 

A magnitude of the methods is inspired by quantum mechanics while not being meant for quantum hardware including \textit{quantum cuts} for image segmentation \cite{Aytekin2014}, \textit{wave kernel signatures}  \cite{shape} for mesh matching and \textit{genetic quantum algorithms} \cite{knapsack} for the knapsack problem. 
None of these methods operates on permutation matrices. 
A related probabilistic physics-based technique which can be applied to problems on graphs is simulated annealing \cite{Kirkpatrick1983}. 
In \cite{herault1990symbolic}, \eqref{DASProblemIntro} arises for 
finding the maximal pair of subgraphs between two graphs. 
The vertices of these graphs represent the interest points in the respective images one wants to match. 

\section{Minimization of Quadratic Objectives over  Permutation Matrices}\label{sec:QGM} 

In this section, we derive three variants of our QGM approach, \textit{i.e.,} baseline variant (Sec~\ref{sec:permutationMatrices}), the variant with row-wise penalties which reduces the penalty-based influence on the problem in overall (Sec.~\ref{ssec:rowwise_penalty}) and the variant with eliminated variables
(Sec.~\ref{sec:PermutationTrick}). 

\label{sec:permutationMatrices} 
\subsection{Baseline Quantum Graph Matching} 
To rephrase \eqref{DASProblemIntro} in the form of \eqref{eq:OptProb}, we have to do two conversions: 1) Switch from $x_{i} \in \{0,1\}$ to $s_i \in \{-1,1\}$, and 2) Switch from the constrained problem over permutation matrices to an unconstrained problem. 
The first one is straightforward by substituting $s_i = 2x_i - 1$, yielding 
\begin{equation}
Q = \frac{1}{4}W, \qquad q = \frac{1}{2} (\mathds{1}^TW+\mathbf{c}), \label{eq:VariableChange}
\end{equation}
between the costs associated with \eqref{DASProblemIntro} and \eqref{eq:OptProb}, respectively, up to a constant (assuming without restriction of generality that $W$ is symmetric, and denoting by $\mathds{1}$ a vector of ones). %

The second conversion is less trivial. 
First of all, note that the set $\mathbb{P}_n$ of permutation matrices can be written as 
\begin{equation} 
\begin{aligned}
    \label{eq:permutationsBeginning}
    \mathbb{P}_n =  \{  X \in \mathbb{R}^{n \times n}~|~ X_{ij} \in \{0,1\},\\ ~ \sum_i X_{ij} = 1~ \forall j,~ \sum_j X_{ij}=1~\forall i \}.
\end{aligned} 
\end{equation} 
Note that -- besides the binary constraint addressed with the above transformation -- $\mathbb{P}_n$ is now characterized by linear equality constraints. Thus, after the vectorization of $X \in \mathbb{P}_n$ to $\mathbf{x}\in \mathbb{R}^{n^2}$, %
one can phrase such constraints in matrix-vector form as $A\mathbf{x}=\mathbf{b}$, for a suitable matrix $A$ and a vector $\mathbf{b}$. 
For $n=2$, as an example, $A$ and $\mathbf{b}$ are given by: 
\begin{equation}
    A= \begin{pmatrix}
1 &1&0&0 \\
0&0&1&1 \\
1&0&1&0 \\
0&1&0&1
    \end{pmatrix}, \quad \mathbf{b} = \begin{pmatrix}
    1\\
    1\\
    1\\
    1
    \end{pmatrix}.
\end{equation}
For higher dimensions, $b$ still contains only ones. The matrix $A$ for arbitrary $n$ can be described with the formula:

\begin{align}
    A= \begin{pmatrix}
 \text{Id} \otimes \mathds{1} ^{\text{T}}  \\
 \mathds{1}^{\text{T}} \otimes  \text{Id}   
    \end{pmatrix},
\end{align}
where $\text{Id}$ is the $n-$dimensional identity matrix.
It is well known (see \textit{e.g.,} \cite{kochenberger}) that linear equality constraints can be incorporated in QUBO by adding a penalty term to the objective, \textit{i.e.,}  
\begin{align}
\label{eq:constrained_to_unconstrained}
\begin{split}
   & \argmin_{\lbrace \mathbf{x}\in \lbrace 0,1\rbrace^{n^2}|\quad A\mathbf{x}=\mathbf{b} \rbrace } \quad  \mathbf{x}^{\text{T}}W \mathbf{x} + \mathbf{c}^{T} \mathbf{x}\\
     = &\argmin_{\mathbf{x}\in \lbrace 0,1\rbrace^{n^2} } \quad  \mathbf{x}^{\text{T}}W \mathbf{x} + \mathbf{c}^{T} \mathbf{x}+ \lambda ||   A\mathbf{x}-\mathbf{b}  ||^2, 
\end{split}
\end{align}
for sufficiently large $\lambda$. For the sake of completeness, we provide a proof of this statement, the specific appearance of $A$ and $\mathbf{b}$, and the lower bound
\begin{equation}
\label{eq:lowerBoundOneLambda}
    \lambda > \lambda^0 := \frac{1}{2} \left( \sum_{i,j= 1}^{N} \abs{W_{i,j}} + \sum_{i= 1}^{N} \abs{c_i} \right) 
\end{equation}
for \eqref{eq:constrained_to_unconstrained} to provably hold in the supplementary material. 
After rewriting the second line of \eqref{eq:constrained_to_unconstrained} to 
\begin{equation}
    \label{eq:OptProbOneLambda}
    \argmin_{\mathbf{x}\in \lbrace0,1\rbrace^{n^2} } \quad  \mathbf{x}^{\text{T}}(W+ \lambda  A^TA) \mathbf{x} + (\mathbf{c}^{T} - 2 \lambda \mathbf{b}^T A  ) \mathbf{x},
\end{equation}
we arrive at a form of the minimization problem \eqref{DASProblemIntro} to which quantum computing is directly applicable. 

Interestingly, applying quantum computing approaches to \eqref{eq:OptProbOneLambda} often does not yield satisfactory results as we will illustrate in more detail in Sec.~\ref{sec:numericalResults}. 
We conjecture that this behavior is largely due to a significant reduction of the \textit{spectral gap} of the coupling matrix $Q+\lambda \frac{1}{4} A^TA$ with increasing $\lambda$ (after the global scale-normalization). 
As discussed in Sec.~\ref{sec:quantumComputing}, the spectral gap  
\textit{i.e.}, the smallest difference between the smallest and the second lowest eigenvalues of the Hamiltonian \eqref{eq:convexCombiHamitonOps} over time, 
has a crucial influence on how slow the transition needs to be
and how likely the system leaves its ground state during the time evolution. 
The penalty in \eqref{eq:OptProbOneLambda} is not the only reformulation of the constrained QAP as an unconstrained QUBO. 
We next derive two alternate formulations (Secs.~\ref{ssec:rowwise_penalty}--\ref{sec:PermutationTrick}) and evaluate them in terms of their resulting spectral gaps as well as behavior on real quantum hardware (Sec.~\ref{sec:numericalResults}). 

\subsection{The Formulation with a Row-Wise Penalty}\label{ssec:rowwise_penalty} 

As large values of $\lambda$ seem to cause problems, we aim at reducing the penalty-based changes of the problem as little as possible by considering row-wise penalty parameters $\lambda_i$ for enforcing each equation $(A \mathbf{x})_i = b_i$ separately, \textit{i.e.,}  
\begin{equation}
    \label{eq:OptProbMultipleLambda}
    \min_{\mathbf{x}\in \lbrace 0,1 \rbrace^{n^2} } \quad  \mathbf{x}^{\text{T}}W \mathbf{x} + \mathbf{c}^{T} \mathbf{x}+ \sum_{i=1}^{2n} \lambda_i ((A\mathbf{x})_i-b_i)^2. 
\end{equation}
While a change of variables from $x$ to $s$ along with the choice $\lambda_i = \lambda > \lambda^0$ according to \eqref{eq:lowerBoundOneLambda} is obviously sufficient and reduces \eqref{eq:OptProbMultipleLambda} to \eqref{eq:OptProbOneLambda}, the challenge is to find small \textit{computable lower bounds} for  $\lambda_i$, for which the unconstrained problem still \textit{provably} coincides with the constrained problem  \eqref{DASProblemIntro}. 
\vspace{-3pt}
\begin{proposition} \label{thm:Reformulation}
The minimizers of \eqref{eq:OptProb} and \eqref{eq:OptProbMultipleLambda} coincide provided that 
\vspace{-3pt}
\begin{equation}
\lambda_{i} > \lambda_i^0 :=D_{\mathcal{J}_i} + \frac{1}{2}  D_{\{1,...,n^2 \} },
\end{equation}
where $\mathcal{J}_i$ denotes the indices that belong to a column or a row enumerated by $i$ in
\eqref{eq:OptProbMultipleLambda}, $D_{\{1,...,n^2 \} } = D_{\mathcal{J}}$ for $\mathcal{J} = \{1,...,n^2 \}$ and
\begin{equation}
   D_{\mathcal{J}}:= \max_{k\in \mathcal{J}}  (\sum_i  \abs{ (W_{k,i }+ W_{i,k}) } + \abs{ W_{k,k}} + \abs{ c_k}).
\end{equation}
\end{proposition} 
\vspace{-5pt}

\subsection{Inserted Formulation with Eliminated Variables} 
\label{sec:PermutationTrick} 
In continuous optimization, linear equality constraints are frequently used to eliminate variables and reduce the complexity of the underlying problem. 
Even in our discrete setting, the sum-to-one constraints of permutation matrices in \eqref{eq:permutationsBeginning} naturally allow expressing $X_{1,j} = 1- \sum_{i=2}^n X_{i,j}$ for all $j$ as well as  $X_{i,1} = 1- \sum_{j=2}^n X_{i,j}$ as long as one ensures that the variables to be replaced remain in $\{0,1\}$. 
In other words: 
\begin{lemma} \label{lm:Permutation}
The set in $\mathbb{P}_n$ of \eqref{eq:permutationsBeginning} can be written as
\begin{align*} 
&\mathbb{P}_n=\lbrace \\
&\resizebox{0.49\textwidth}{!}{ $\left( \begin{array}{rrrrr}
2 - n+\sum_{i,j= 2}^n  x_{i,j}  & 1 -\sum_{i= 2}^n  x_{i,2} & \dots & 1 -\sum_{i= 2}^n  x_{i,n} \\
1- \sum_{i= 2}^n x_{2,i} & x_{2,2} & ... &  x_{2,n} \\
\vdots & \vdots & \ddots & \vdots \\
1- \sum_{i= 2}^n x_{n,i} & x_{n,2} & ... &  x_{n,n} 
\end{array}\right)$} \\
& \in \mathbb{R}^{n\times n}  | \forall i \in \lbrace 2,...,n \rbrace , j \in \lbrace 2,...,n \rbrace \text{ }  x_{i,j} \in \lbrace 0,1 \rbrace, \\ & \sum_{i,j= 2}^n  x_{i,j} \in \lbrace n-2, n-1 \rbrace,  \quad \\ & \forall j,i,k \in \lbrace 2,...,n \rbrace, \;\; i\neq k, \;\; x_{j,i} x_{j,k} =0 \wedge     x_{i,j} x_{k,j} =0    \rbrace. 
\end{align*}
\end{lemma}
This reformulation of the constraint set allows us to reduce the amount of needed qubits from $n^2$ to $\left(n-1 \right)^2$ by incorporating all remaining constraints via penalties as stated in the following proposition. 

\begin{proposition} \label{thm:ReformulationPermutation}
The problem \eqref{DASProblemIntro} has the same solution as the following QUBO for sufficiently large $\lambda_{1}$ and $\lambda_2$: 
\begin{align*}
&\min_{\mathbf{x}\in \lbrace 0 ,1\rbrace^{(n-1)^2} } \quad  \mathbf{x}^{\text{T}} \tilde{W} \mathbf{x} + \tilde{\mathbf{c}}^{T} \mathbf{x} + \lambda_1 \sum_{j,k=0 \quad j\neq k}^{(n-1)^2} \chi(j,k) x_j x_k \\&
+ \lambda_2 \left( \sum_{i=1 }^{(n-1)^2}x_i -(n-1)  \right) \left( \sum_{i=1 }^{(n-1)^2}x_i -(n-2) \right) \\
&\chi(j,k) := \begin{cases} 1 \quad &\text{ for } \lfloor \frac{j}{n-1} \rfloor =\lfloor \frac{k}{n-1} \rfloor  \\
\quad &\vee   j-(n-1) \lfloor \frac{j}{n-1}\rfloor= k-(n-1) \lfloor \frac{k}{n-1} \rfloor \\
0 \quad &\text{otherwise}
\end{cases},
\end{align*}
where the specific definitions of the reduced $(n-1) \times (n-1)$ matrix $\tilde{W}$, the vector $\tilde{\mathbf{c}} \in \mathbb{R}^{n-1}$ and the precise lower bounds for $\lambda_1$ and $\lambda_2$ are provided in the supplement. 
\end{proposition}

The reduction in the number of required qubits is a clear advantage of the above \textit{inserted} formulation, especially in the current early stage of AQC-ers, when 
the number of supported logical qubits is limited  \cite{DWAVEQPU2020}. 
On the other hand, the asymptotic difference between the $n^2$ and $(n-1)^2$ (of the inserted formulation) vanishes, such that their spectral gaps, and, more importantly, their performance on real AQC-ers are the deciding factors we evaluate in the next section. 

\section{Experimental Results} 
\label{sec:numericalResults} 
We next test our methods for quantum graph matching on random problem instances in Sec.~\ref{ssec:random_instances} and real data in Sec.~\ref{ssec:real_experiment}.
We also analyze in Sec.~\ref{ssec:range_couplings} the influence of experimental errors in the coupling parameters to the measured results. 
All in all, we investigate the following questions: 
\begin{itemize}[leftmargin=*] 
    \setlength{\itemsep}{0pt} 
    \item Do the three equivalent reformulations of \eqref{DASProblemIntro}, \textit{i.e.,} the baseline, the row-wise penalty and the inserted formulations lead to different spectral gaps? 
     \item Do larger spectral gaps improve AQC-ing results? 
    \item Do quantum computing approaches provide advantages over classical relaxation methods for solving \eqref{DASProblemIntro}? 
    \item Do AQC-ers -- as well as their simulations, \textit{i.e.,} numerical solutions of \eqref{eq:schroedinger} --  solve our problem? 
\end{itemize} 

We prepare the weight matrix $W$ on a classical computer, and access D-Wave 2000Q remotely in the cloud via \textit{Ocean Tools Leap 2} \cite{DWAVE_LEAP}. 
In all experiments, we report the solution distributions and analyse the probability and the energy of the most frequent solution. 
The ground-truth permutations are calculated by brute force and  compared qualitatively with the expected outcomes on real data. 
Minor embedding of QGM results in a fully connected logical graph of qubits. 
For $n=3$, the number of physical qubits in the minor embedding amounts to ${\approx}27$. 
\subsection{Evaluation on Random Problem Instances} 
\label{ssec:random_instances} 
\begin{figure}[t]
    \centering
  \includegraphics[width=\linewidth]{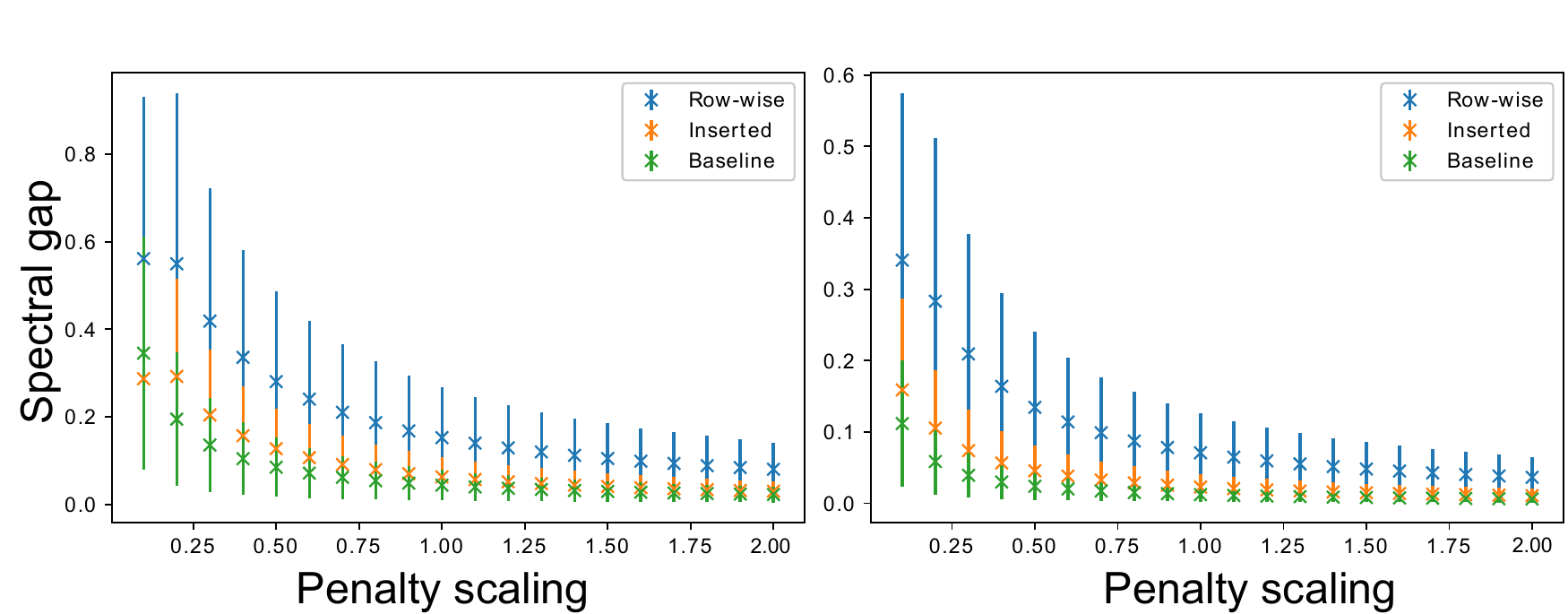}
    \caption{Illustrating the spectral gap for our baseline, %
    the row-wise penalty and inserted formulation as a function of the global scaling of the penalty term for $n=3$ (left) and $n=4$ (right). As we see, the row-wise penalty leads to the largest spectral gap which is favorable for AQC-ing. %
    }
    \label{fig:spectralGap}
\end{figure}

As the first step, we create random problems of the form \eqref{DASProblemIntro} by drawing all entries in $W$ and $c$ uniformly from $[ -1,1]$. %
Those are representative for a broad class of problems arising in 3D vision such as mesh and point cloud matching under isometry, and rigid point set alignment. 
 We compute the spectral gap for each of our three formulations via an iterative Lanczos method  \cite{Lanczos50aniteration}, since this calculation requires eigenvalue decomposition of large quadratic matrices with the dimension $2^{n^2}$. 
Fig.~\ref{fig:spectralGap} shows the spectral gaps of all three formulations averaged over ten different problem instances in dimensions $n=3$ and $n=4$, when the strength of the overall penalty is varied by a global scale factor (x-axis). 
A scale factor of $1$ refers to the setting where the constrained and unconstrained problems are provably equivalent according to our derivations in Sec.~\ref{sec:permutationMatrices}. 
As we see, the spectral gap reduces significantly for all formulations and instances with the increasing strength of the constraint penalty. 
Among the three formulations, the row-wise penalty consistently results in the largest spectral gap which we expect to be favorable for real and simulated AQC-ing. 

\begin{figure}[t]
    \centering
  \includegraphics[width=\linewidth]{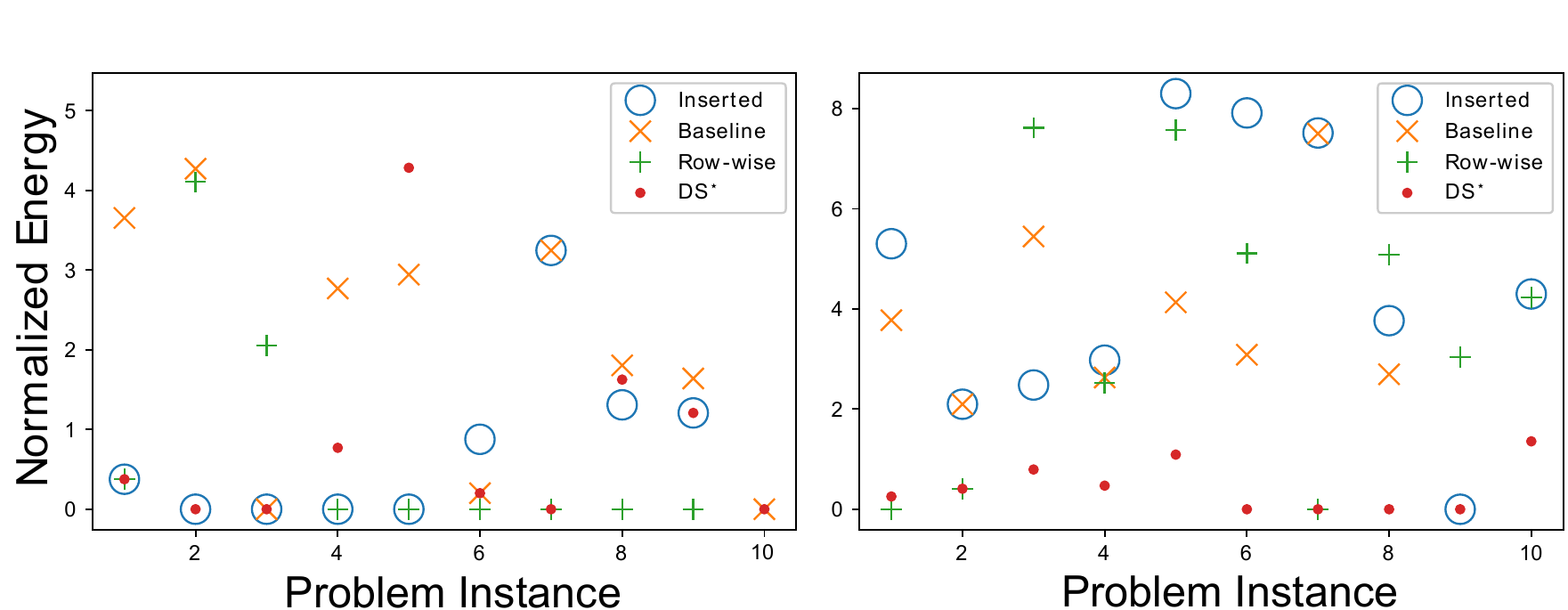}
    \caption{Comparing quantum computing on DWave 2000Q for our three reformulations and the relaxation-based method DS$^*$ \cite{relaxations} for computing $3\times 3$ (left) and $4 \times 4$ permutations (right) on ten random problem instances. 
    }
    \label{fig:resultsRandom}
\end{figure}

Fig.~\ref{fig:resultsRandom} shows the different results of our three reformulations run on D-Wave, and the recent relaxation-based algorithm \cite{relaxations} for $n \in \{3,4\}$ over ten instances of \eqref{DASProblemIntro}. 
For each instance, we compute the global optimizer by brute force and normalize (shift) the energy by $-f_{opt}$ for each method. 

For the AQC-ing approach, we use $500$ annealing cycles on the D-Wave 2000Q and select the most frequent solution among all annealing cycles as our final solution. 
On D-Wave, logical problem qubits are mapped to multiple physical qubits by chaining. 
For $n=4$, we set the chain strength as the 
maximal element of the $q$ vector in \eqref{eq:VariableChange}. This leads to an automatic adaption of the chain strength to the problem instance. For $n=3$, this does not lead to further improvements over 
empirical fixed values of the chain strength. 
Here, we have the chain strengths $20,30$ and $35$ for the inserted, the row-wise and the baseline method,  respectively.
Up to a $100\mu s$ break, we insert in the middle of the annealing path, the annealing time $\tau $ is $80\mu s$ for $n=4$. 
For $n=3$, $\tau = 20\mu s $ already yields desired result. 

The simulation with QuTiP \cite{qutip} showed that increasing the $\tau$ yields a narrower distribution. 
For $n=3$, the simulation can find the correct optimum as the highest peak regardless of the method variant. 
The situation changes for $n=4$ 
and the simulation with the \textit{baseline} formulation does not yield satisfactory results. 
The method with inserted constraints, on the other hand, finds the minimum every time. 

\newcolumntype{P}[1]{>{\centering\arraybackslash}p{#1}}

\begin{table}[t]
\footnotesize
\begin{tabular}{|c|c|c|c|c|P{1.3cm}|}
\hline
$\boldsymbol{n}$ & \textbf{inserted}  & \textbf{baseline} &  \textbf{row-wise} & $\textbf{DS}\boldsymbol{^{\star}}$  \cite{relaxations}& \textbf{worst permutation} \\ \hline \hline 
3 & \multicolumn{1}{r|}{$1.49$} & \multicolumn{1}{r|}{$2.12$} & \multicolumn{1}{r|}{$2.27$} & \multicolumn{1}{r|}{$0.85$} &\multicolumn{1}{r|}{$4.62$} \\ \hline
4 &\multicolumn{1}{r|}{$5.68$} & \multicolumn{1}{r|}{$7.37$} & \multicolumn{1}{r|}{$7.40$} & \multicolumn{1}{r|}{$0.43$}& \multicolumn{1}{r|}{$10.23$} \\ \hline
\end{tabular}
\caption{Average energies for $n=3,4$.}
\label{tab:summary}
\end{table}

Table \ref{tab:summary} summarizes the mean normalized  energies over ten instances for $n=3$ and $n=4$, with 
the averages computed so that $0$ is the best achievable value. 
Note that if AQC-ing yields a state that is not a valid permutation, the worst permutation is used for averaging. 
Compared to the simulation, where we were always able to find the global solution with the inserted or the row-wise method for large enough $\tau$, AQC-ing on quantum hardware is significantly more difficult. 
As we can see in Fig.~\ref{fig:resultsRandom}-(left), the most probable states have the lowest energy for the inserted and the row-wise method, and the results are comparable to DS$^{\star}$ \cite{relaxations} in $n=3$. 
For $n=4$ the DS$^{\star}$ performs clearly the best. 
Interestingly, for $n=3$, the row-wise penalties were able to
substantiate 
their advantages with respect to the spectral gap in practice, yielding several problem instances on which it outperformed DS$^*$ and being on par with it on average. 
While the dimension of the underlying problems is of course small (for $n=3$, there only exist six different permutation matrices), we consider these results to be very promising, as they illustrate the potential of AQC-ing beyond pure simulations with future generation of AQC-ers. 
\subsection{Ablative Study on the Coupling Parameters} 
\label{ssec:range_couplings} 

\begin{figure}[t]
    \centering
\includegraphics[width=0.49 \textwidth]{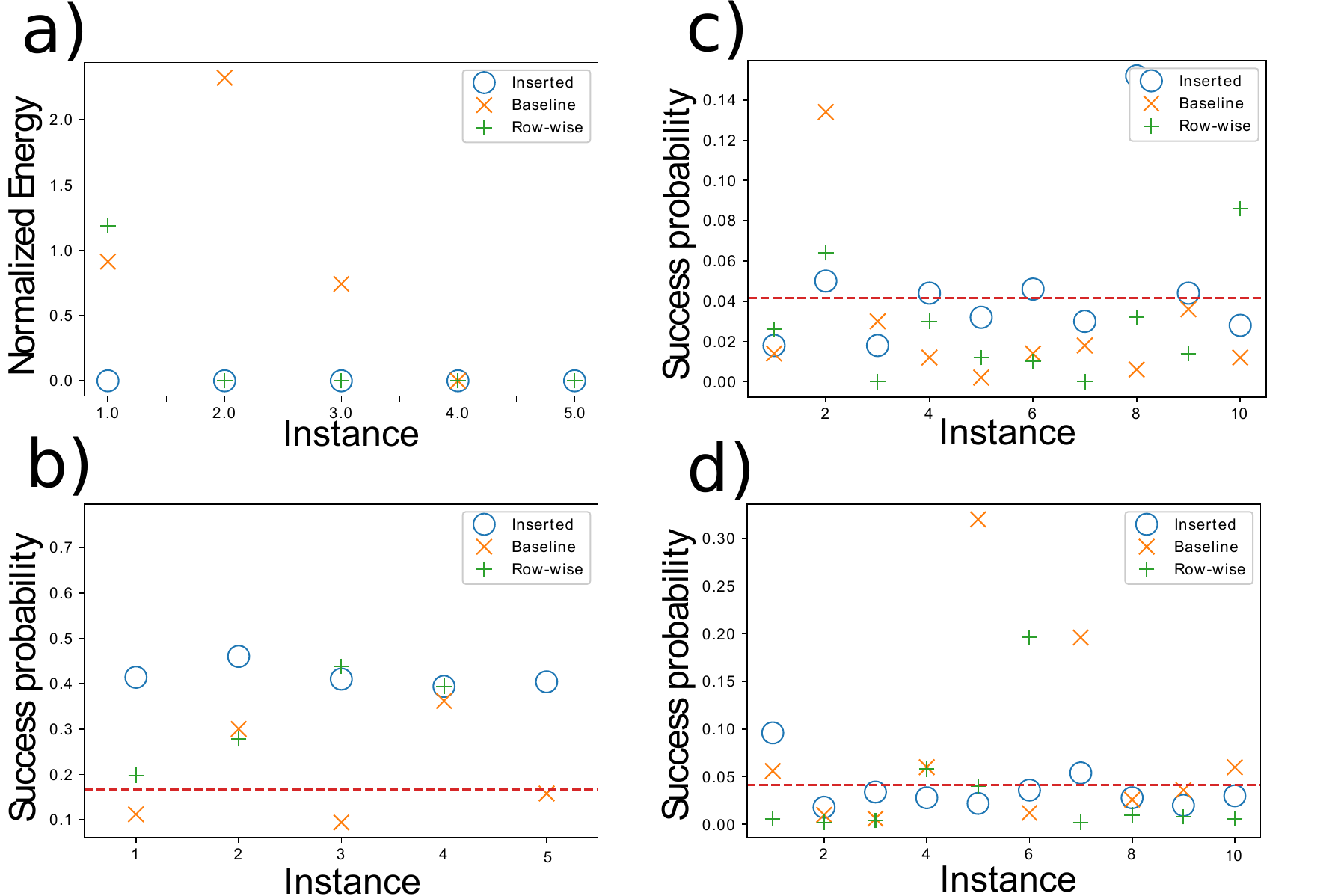}
    \caption{Problem instances, where multiple elements are randomly chosen to be zero. In \textbf{a)} and \textbf{b)}, the problem instances are for $n=3$. 
    Here, less than half of the entries in $W$ and $c$ are randomly set zero. 
    \textbf{c)} and \textbf{d)} show the success probability for $n=4$.
    For \textbf{c)}, exactly half of the entries are chosen zero and for \textbf{d)}, exactly $3/4$ are chosen zero. }
    \label{fig:resultsRandomSparse}
\end{figure}

On D-Wave, the input $Q$ and $\mathbf{q}$ values are scaled so that $-1<Q_{i,j}<1 $ and $-2<q_i<2$. 
Additionally, there are multiple further error sources for the couplings and biases \cite{DWaveError}. 
Therefore, it can happen that the input $Q_{i,j}= 0.5$ is realized as $0.51$, for example. 
As we saw in Sec.~\ref{sec:permutationMatrices}, Eqs.~\eqref{eq:constrained_to_unconstrained} and \eqref{eq:OptProbMultipleLambda}, the Hamiltonian has a part that depends on the problem instance $W, c$ directly and includes the regularization term which only depends on $W, c$ in the penalty parameters. The regularization part yields the same energy for every valid permutation.
Let $Q_{\text{reg}}$ and $q_{\text{reg}}$ be the couplings and biases that stem from the regularization term.
If permutations only marginally differ in the qubit couplings, it is difficult to distinguish between them during a quantum anneal. 
This suggests that we have to choose the penalties as small as possible while already delimiting from random binary strings. 
Assuming that all values in $W$ are of the same order of magnitude, 
$\lambda$ would be $n^4$ times bigger than $W$ values 
for the baseline method. 
Our calculations also confirm that for $n=4$, the maximal value of $c$ that is due to the penalty, is $500$ times larger than the maximal value of $c$ in the problem itself. 
One easy way to have a larger (more beneficial) ratio between the range of the  original $W$ values and the range value in the penalty term is to consider problems 
with $W$ entries randomly set to zero. 
The connectivity should stay the same, due to the regularization. 
For $n=4$, we either set a half or $\frac{3}{4}$ of the entries of the problem instance to zero. 
The results in Fig.~\ref{fig:resultsRandomSparse} show that for the case $n=4$, random guessing is still better most of the time. 

\subsection{Experiments with Real Data} 
\label{ssec:real_experiment} 
We now solve several real problems of matching two graphs originating from two 3D shapes deformed in a near-isometric way. 
Assuming that we have an edge-weight or a notion of distance $d_1(v^1_i, v^1_k)$ between nodes $v^1_i$ and $v^1_k$ of our first graph, and, similarly, a distance $d_2(v^2_j, v^2_l)$ between nodes $v^2_j$ and $v^2_l$ of our second graph, a common choice for the quadratic costs in \eqref{DASProblemIntro} is 
\begin{equation}
    \label{eq:isometricMatching}
 \sum_{i,j,k,l} X_{i,j}  X_{k,l} |d_1(v^1_i, v^1_k)-d_2(v^2_j, v^2_l)|. 
 \end{equation}
The above formulation assures that if $v^1_i$ is matched to  $v^2_j$ and $v^1_k$ is matched to  $v^2_l$, then the distance between $v^1_i$ and $v^1_k$ is similar to the distance between $v^2_j$ and $v^2_l$. 

\begin{figure}[t]
    \centering
   \includegraphics[width=0.47\textwidth]{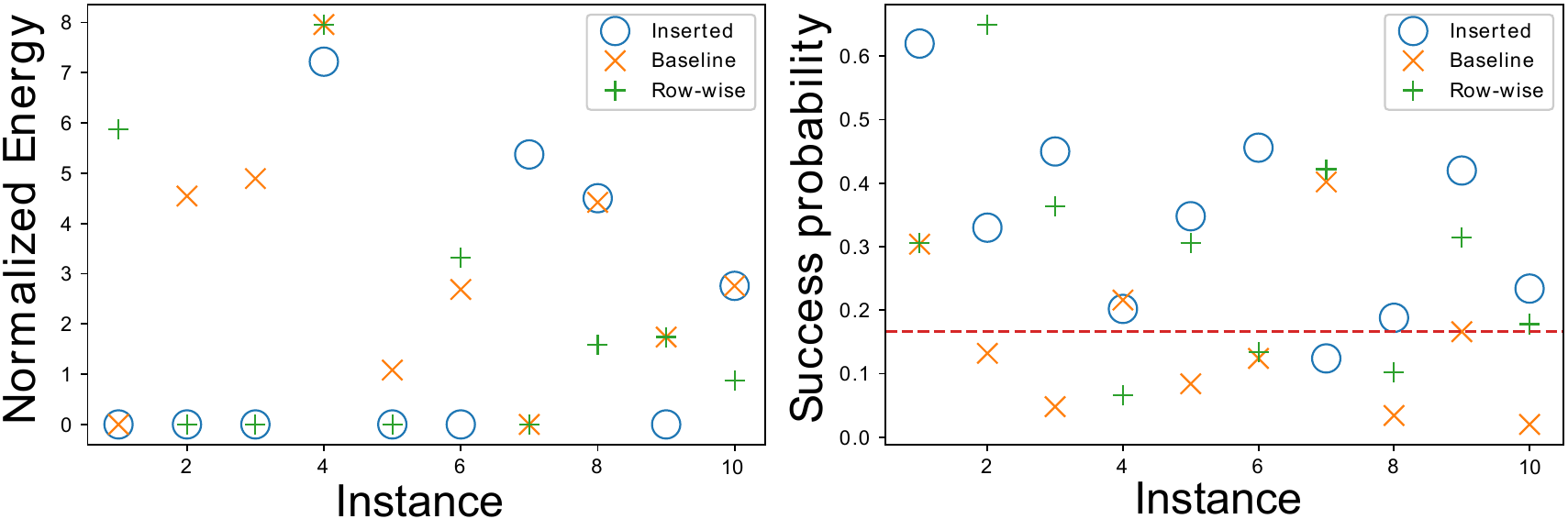}
    \caption{Performance of the quantum annealer in the shape matching problem for $n=3$. 
    } 
    \label{fig:shapeMatchingQuant}
\end{figure}

We consider different instances of \eqref{eq:isometricMatching}: In our first shape matching example, we aim to identify the corresponding points on two differently deformed instances of the same shape. 
In this case, $d_1$ and $d_2$ denote the geodesic distance on the respective shapes, and we introduce an additional linear term based on the Euclidean distance of the points after an initial rigid registration following \cite{ChenandKoltun}. 
Fig.~\ref{fig:shapeMatching} visualizes the matching of the two shapes along with the histogram we obtained from $500$ runs of this problem instance on the D-Wave 2000Q. 
See Fig.~\ref{fig:shapeMatchingQuant} for the performance comparison of different QGM variants. 
We observe that the results with real 3D shapes match coarsely the distributions of solutions to experiments with random instances in Sec.~\ref{ssec:random_instances}. 

\begin{figure}[t]
    \centering
  \includegraphics[width=0.47\textwidth]{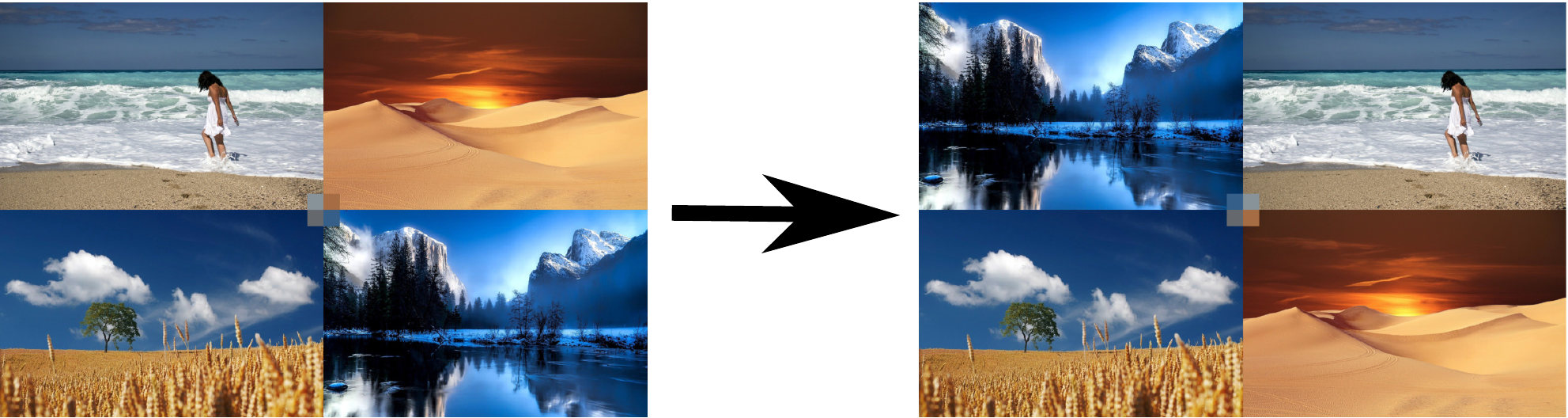}
    \caption{Example of image matching for $n=4$. 
    } 
    \label{fig:imageSorting} 
\end{figure}

\begin{figure}[t]
    \centering
  \includegraphics[width=0.47\textwidth]{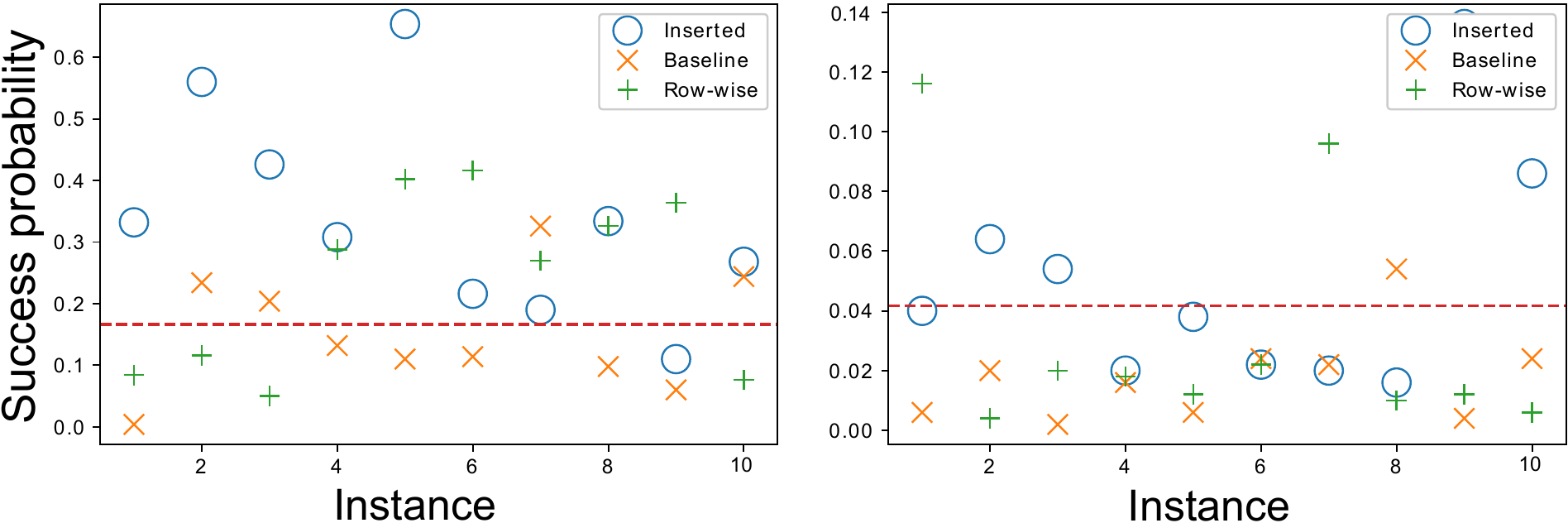} 
    \caption{The probability to measure the optimum in one run  compared to random guessing of permutations for $n=3$ (left) and $n=4$ (right). The problem instances are dense, \textit{without} zero entries. 
    } 
    \label{fig:probabilityRandom} 
\end{figure}

Fig.~\ref{fig:imageSorting} illustrates another application of costs \eqref{eq:isometricMatching} for sorting images according to their mean color as considered in \cite{Dym2017}, see Fig.~\ref{fig:imageSorting} for the visualization. 
In this setting, $d_1$ represents the difference in the mean image color, and $d_2$ is the Euclidean distance on the $2 \times 2$ image grid. 
The mean image color is sampled from the middle region of each image. 
As we can see, after the sorting on the right, the images are ordered from predominant blue on the top left (mountains with a lake), to the predominant yellow (desert with the sunset). 
This problem, likewise, has multiple possible minimizers, and the quantum annealer finds the optimum more often than the other (non-optimal) states.

\section{Further Observations and Limitations} 
\label{sec:discussion}
We were able to successfully confirm in our experiments the validity of our QGM approach for $n=3$. 
On a wide range of randomly generated problem instances, our method outputs optimal valid permutations with high probability. 
On the current D-Wave hardware, the results for $n=4$ are significantly worse, \textit{i.e.,} neither do we obtain a conclusive best choice between the three different formulations, nor do we get near the performance of the DS$^*$ heuristic \cite{relaxations}. 
Fig.~\ref{fig:probabilityRandom} shows the probabilities of obtaining the optimal solution based on $500$ anneals per instance. 
For most instances, the different formulations performed even worse than random guessing, because they can also output vectors that do not correspond to permutations. 
These results cannot be due to chain breaks, because the D-Wave compiler did not report any. 
Another interesting quantity is the length of the chain. 
For $n=2$, our chains have maximal length of two, 
whereas for $n=3$, there are also chains of length three and four. For $n=4$ the chains have length five and six. 

\noindent\textbf{Remark on IBM Q Experience.} 
We also run tests via the IBM Q Experience \cite{IBMQExp}. 
They offer not an AQC-er but a circuit-based universal quantum computer, where the algorithm can be viewed as a circuit consisting of so-called \textit{quantum gates}. 
To test our method on it, 
we discretize our Hamiltonian with constant time steps and apply trotterization \cite{nielsen} for the decomposition into quantum gates. 
For higher-dimensional problems, the number of gates rises, which leads to errors. 
We thus are only able to successfully realize an adiabatic transition for two qubits. 
By inserting two equality restrictions, we can optimize over $2\times2$ permutations. 
We provide more details on this alternative direction in the supplementary material. 
\section{Conclusions}\label{sec:conclusions} 
The quadratic assignment problem which we tackle in this paper is a hard problem to be mapped and solved on a real quantum annealer, and we make a further successful step to solve it on real AQC-ers. 
The difficulties come from the high demand in connectivity between the qubits and the strong scaling in the qubit number ($n \log(n)$ in the best case). 
We achieve expected results for $3\times 3$ permutations and demonstrate the potential of the new quantum method for solving quadratic assignment problems in 3D vision, whereas our formulation can be applied to problems in different dimensions which is demonstrated experimentally. 
The spectral gap analysis shows that it greatly matters how the problem is mapped to a QUBO and subsequently solved by an AQC-er. 
We expect a significant impact of the future quantum technology on 3D v11ision, while the accessibility of modern AQC-ers for research purposes allows us to lay the foundation  already today. 
In future work, we will
investigate other structures such as point clouds with outliers, more nodes and partial matches. 

\noindent\textbf{Supplementary Material.} 
Our supplement contains the proofs of Lemma  \ref{lm:Permutation} and Prop.~\ref{thm:Reformulation} as well as more experimental results.%

\noindent\textbf{Acknowledgement.} 
This work was partially supported by the ERC Consolidator Grant 4DReply (770784). 

{\small
\bibliographystyle{ieee}
\bibliography{egbib}
}

\onecolumn
\setcounter{section}{0}
\renewcommand\thesection{\Alph{section}}
\newcommand{\suppsection}{\subsection}
\clearpage
\begin{center}
\textbf{\Large Adiabatic Quantum Graph Matching with Permutation Matrix Constraints}\\
\vspace{3pt}
\textbf{\Large---Supplementary Material---}
\end{center}
\makeatletter

\input{3DVSupplementary.tex}

\end{document}

%% file: quantum_computing_new.tex
\section{Modern Adiabatic Quantum Computing}  \label{sec:quantumComputing} 

We next review the basics of AQC-ing. 
We suppose that the reader is familiar with general concepts of linear and tensor algebra, calculus as well as operator theory, and 
no preliminary knowledge of quantum physics is assumed. 

\noindent\textbf{State Space of Quantum Systems.} 
While the state of a quantum-mechanical system is generally described by a normalized vector in a Hilbert-space $\mathcal{H}$, a simple example for a quantum mechanical property is the spin of electrons, in which case the Hilbert-space $\mathcal{H}$ is the two-dimensional complex Euclidean space, $\mathcal{H} = \mathbb{C}^2$. 
The corresponding quantum state, \textit{i.e.,} an element of the Hilbert space with norm $1$, commonly denoted as $\ket{\psi} \in \mathbb{C}^2$, is called a \textit{qubit}. 
For vectors $\ket{0}, \ket{1}$\footnote{we use Dirac notation; $\ket{0}$ and $\ket{1}$ represent an orthonormal basis} that form an orthonormal basis of $\mathbb{C}^2$, any such quantum state can be represented as a linear combination of the basis vectors, \textit{i.e.,} 
\begin{equation}
\label{eq:super}
\ket{\psi} = \alpha \ket{0} + \beta \ket{1}, \;\, \alpha, \beta \in \mathbb{C}, \;\, \abs{ \alpha }^2 + \abs{ \beta} ^2 =1. 
\end{equation}
Quantum mechanics dictates that measuring a qubit in the basis  $\ket{0},\ket{1}$ 
yields the states $\ket{0}$ and $\ket{1}$ with the probability  $\abs{\alpha}^2$ and $\abs{\beta}^2$, respectively. 
%
The measurement itself influences the state such that it collapses to either $\ket{\psi}=\ket{0}$ or $\ket{\psi} = \ket{1}$. 
The state with $\abs{\alpha}^2 = \abs{\beta}^2 = 0.5$ is denoted by $\ket{+}$. 
The state $\ket{\psi}$ for $\alpha, \beta \neq 0$ is called a \textit{superposition}. 

\noindent\textbf{Composite Quantum Systems.} 
Currently, a common approach to realize a quantum computer is to build experiments with multiple qubits, in which case the Hilbert space for representing the overall system is expressed as the tensor product of the individual Hilbert spaces. 
 If one now thinks of multiple spin-$\frac{1}{2}$ particles or $n$ qubits, we obtain states $\ket{\psi} \in \mathcal{H} \left(\mathbb{C}^2\right)^{ \otimes n}:=\mathbb{C}^2 \otimes ...\otimes \mathbb{C}^2  $. Note that if the spins of the electrons were completely independent from each other, we would only need $2n$ many parameters to describe a state. The reason we need the $2^n$ parameters for the state space is that for so called \textit{entangled states} the spins of the individual systems cannot be treated separately.

\noindent\textbf{Time Evolution of Quantum Systems.} 
The time evolution of a quantum system, \textit{i.e.,} description how a quantum state $\ket{\psi}$ changes over time under an external influence, is given by the Schr$\ddot{\text{o}}$dinger equation: 
\begin{equation}
\label{eq:schroedinger}
i \hbar \frac{\partial}{\partial t} \ket{\psi(t)} = H(t)\ket{\psi(t)},
\end{equation}
where $i$ is the imaginary unit, $\hbar$ is the reduced Planck constant, $t$ is the time and $H(t)$ is the so-called \textit{Hamiltonian}, which is a self-adjoint, linear operator on the Hilbert space $\mathcal{H}$ characterizing the energy of the physical system. 
If the Hamilton operator $H$ in \eqref{eq:schroedinger} is not time-dependent, and the initial state $\ket{\psi(0)}$ is an eigenvector of $H$ to the eigenvalue $\lambda \in \mathbb{R}$, \eqref{eq:schroedinger} has the close-form solution
\begin{equation}
    \ket{\psi(t)} = \exp{-\frac{i}{\hbar} \frac{t}{\lambda}} \ket{\psi(0)},
\end{equation}
such that $\ket{\psi(t)}$ remains in the same state modulo a phase. 

\noindent\textbf{Adiabatic Quantum Computing.} 
AQC solves quadratic unconstrained binary optimization problems (QUBO) on quantum hardware relying on the \textit{adiabatic quantum theorem} \cite{BornFock1928}. 
Assume one faces the following QUBO: 
\begin{equation}
\argmin_{\mathbf{s} \in \lbrace -1,1 \rbrace ^n} \quad   \mathbf{s}^T Q \mathbf{s} + \mathbf{q}^T \mathbf{s},
\label{eq:OptProb}
\end{equation}
with an $n\times n$-matrix $Q$ and an $n$-dimensional vector $\mathbf{q}$. 
Furthermore, assume one is able to set up a quantum system with a Hamilton operator $H(0)=:H_B$ in a state $\ket{\psi(0)}$ that is an eigenvector to the smallest eigenvalue of $H_B$, and one constructs a second Hamilton operator $H_P$ whose eigenvector to the smallest eigenvalue is a solution to \eqref{eq:OptProb}. 
Then, \cite{BornFock1928} dictates that the time evolution of \eqref{eq:schroedinger} with 
\begin{equation}
\label{eq:convexCombiHamitonOps}
H(t)= \frac{t}{\tau} H_{P} + \Big( 1-\frac{t}{\tau} \Big) H_B, 
\end{equation}
for $t\in [0,\tau]$ with large $\tau$ to slowly transition between $H_B$ and $H_P$, remains in the eigenvector with the smallest eigenvalue (provided that it remains unique and non-degenerate), so that measuring the corresponding state $\ket{\psi(\tau)}$ at $t = \tau$ yields the solution to \eqref{eq:OptProb}. 
For the cases we investigate in the paper, $H_P$ is a diagonal matrix and for $\ket{\psi(0)}$, all components are equal in the computational basis. 
To experimentally realize an evolution with the Hamiltonian $H_P$, 
one needs to provide couplings between the qubits in form of the matrix $Q$ and the biases as the vector $q$ upon \eqref{eq:OptProb}. 
Of course, an infinitely slow transition between the two Hamilton operators, \textit{i.e.,} $\tau \rightarrow \infty$, is impossible (and would eliminate any computational advantages). 
Luckily, it has been shown that suitable finite choices of $\tau$ to remain in the adiabatic case depend on the \textit{spectral gap} of the problem, \textit{i.e.,} the difference between the smallest and second lowest eigenvalue. 
For more details on how to estimate $\tau$ and spectral gap analysis, see \cite{consistency,jarret2014adiabatic}. 

\noindent\textbf{D-Wave 2000Q.} 
D-Wave 2000Q is an AQC-er with $2048$ qubits arranged on a \textit{Chimera} graph, \textit{i.e.,} a $16 \times 16$ set of  cells with eight qubits each. 
The qubits inside a cell are fully connected, whereas there are only eight connections to qubits from two other neighbouring cells \cite{DWAVEQPU2020}. 
The computation always starts with the initialization state $ \ket{\psi(0)} = \ket{+}^{\otimes n}$, where $n$ is the number of physical qubits in the problem mapping to the Chimera graph, also known as \textit{minor embedding}. 
Logical problem qubits are often mapped to chains of physical qubits, due to the restriction on the physical qubits connectivity. 
The D-Wave compiler performs the minor embedding automatically, and the programmer loads couplings $Q$ and biases $q$ for the target problem. 
Moreover, additional settings can be provided such as the chain strength, annealing time and the annealing path. 

The default duration of an anneal is $20 \mu s$. 
To solve a problem, usually multiple anneals of the same problem instance are performed, since there is no guarantee (due to the physical characteristics of current quantum hardware) that D-Wave will return optimal solutions. 

The couplers are realized with loops of niobium. 
During the annealing, the processor imposes magnetic fluxes on the couplers corresponding to the provided $Q$ matrix. 

%% file: 3DVSupplementary.tex
This supplementary material provides details on the derivations of the Quantum Graph Matching (QGM) approach as well as more experimental results. 
The proofs of Lemma \eqref{lm:Permutation} and Prop.~\eqref{thm:Reformulation} of the main matter can be found in Sec.~\ref{sec:proofs}. 
Sec.~\ref{sec:IBM_Q_One} complements the description of our experiments with the quantum computers, accessible via IBM Quantum Experience. 
Next, we show additional results of QGM for point cloud alignment in Sec.~\ref{sec:point_cloud_matching} and nearly isometric non-rigid shape registration in Sec.~\ref{sec:near_isometric_matching}, followed by 
a discussion of the minor embedding and negative results in Secs.~\ref{sec:minor_embeddings} and \ref{sec:discussion_negative_results}, respectively. 
Note that D-Wave provides one minute of QPU time on their annealers for research purposes per month. 
We obtain three minutes in total during the current project and use $80\%$ of this time for our experiments. 
One minute of time on the quantum annealer can suffice to  run hundreds of problems.

\section{Proofs for Section \ref{sec:QGM}}\label{sec:proofs}

\input{theory}

\section{Experiments on IBM Burlington}\label{sec:IBM_Q_One} 
\begin{figure}[ht] 
\hspace{20pt}\includegraphics[width=0.9\linewidth]{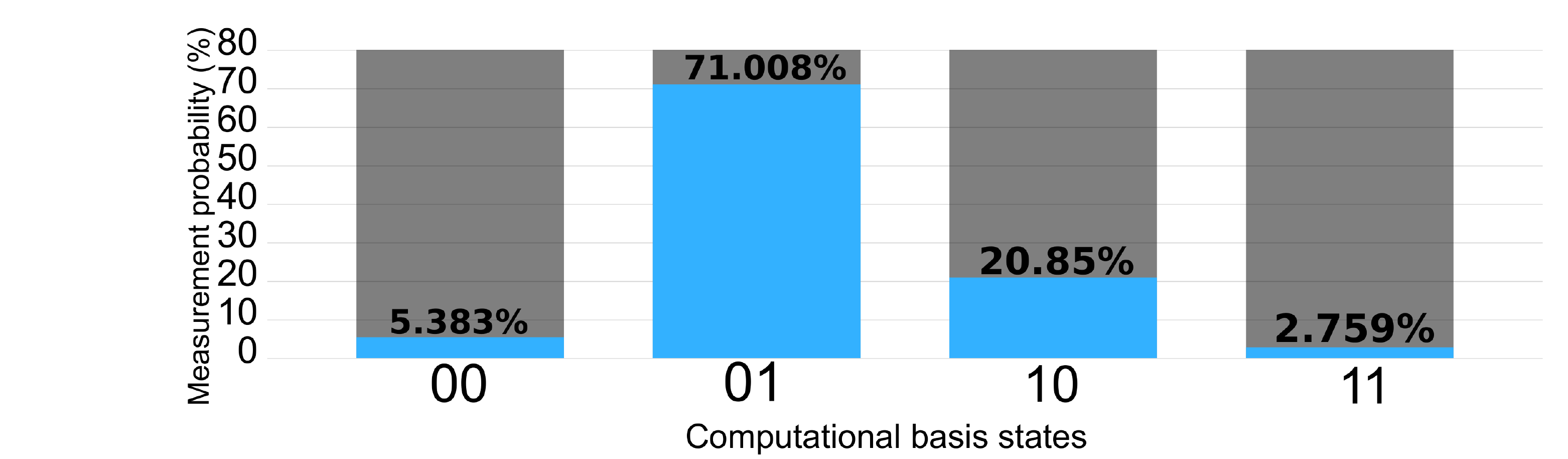} 
\caption{
QGM execution histogram over $8192$ runs using trotterization on a five qubit processor from IBM Quantum Experience. 
The states $\ket{00}$ and $\ket{11}$ are suppressed, because of the proposed regularization terms, \textit{i.e.,} since these states violate that the sum of all elements in a column of a permutation matrix equals to $1$. 
} 
\label{fig:Histogramm} 
\end{figure}

Several of our experiments are conducted on the IBM five qubit QC-er in Burlington \cite{IBMQExp}, which is accessible per cloud. 
In contrast to D-Wave quantum annealers, the circuit-model is used for the design of this quantum computer. 
From a purely theoretical perspective there is some  equivalence \cite{adiabaticequivgate} between adiabatic quantum computing and quantum computing with the circuit model, though due to the technical challenges, 
significant differences remain in practice.
One way to execute a time-dependent Hamiltonian $H(t)$ \eqref{eq:convexCombiHamitonOps} on the IBM machine is to approximate it with $L$ constant Hamiltonians. 
The Hamiltonians, which are constant in time, can be decomposed and expressed with quantum gates via \textit{trotterization} \cite{nielsen}. 
More specifically, choosing the value of the parameter $L$ in the piecewise constant approximation of the Hamiltonian is crucial as we loose the (pseudo-)adiabacity for small $L$, but -- as each gate has a certain chance of introducing an error and a long execution time of the circuit makes errors due to decoherence more probable -- large $L$ are also prone to fail. 
Because of the above-mentioned reasons, we are able to perform the adiabatic evolution for only two qubits on the 5 qubit processor until now. 
Fortunately, using the trick from Sec.~\ref{sec:PermutationTrick}, we can optimise over the $2\times 2$ permutation matrices. 
In our experiment, we performed $50$ time steps. 
We choose the evolution time $0.1$ and use Suzuki expansion of order 2, similar to Kraus \textit{et al.}~\cite{Kraus}. 
The matrix $W$ and the vector $c$ were generated randomly for every of the $20$ iterations. 
Every circuit was executed $8192$ times, which is the maximal amount. 
An exemplary result of an execution is summarized in the histogram in  Fig.~\ref{fig:Histogramm}. %
Although we obtain {slightly different} distributions {each time}, the highest peak always coincides with the second column of the optimal permutation, which is $\ket{01}$ here. 

The recent publication \cite{cao2020speedup} about adiabatic quantum computing on machines from IBM Quantum Experience proposes catalyst Hamiltonian, which could improve the results in future and make it possible to succeed in higher dimensions. 

\section{Point Cloud Matching Example}\label{sec:point_cloud_matching} 
\begin{figure*}[htb]
    \centering
    \includegraphics[width = 1\textwidth]{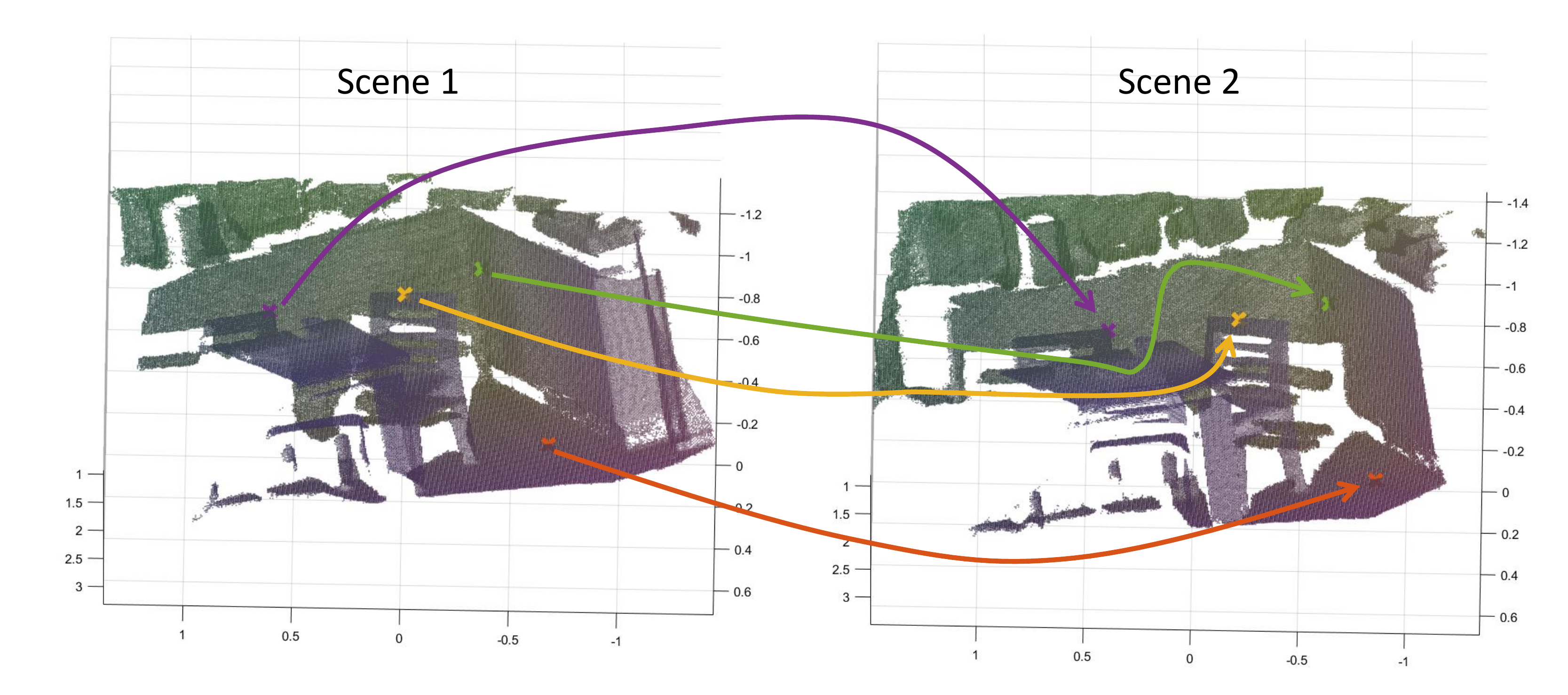}
    \caption{Illustrating the correct  matching between four selected keypoints in two frames of the 7-Scenes 'Redkitchen' dataset, as provided at \url{https://3dmatch.cs.princeton.edu/}}
    \label{fig:pointCloudMatching}
\end{figure*}

To illustrate another application of the studied matching problems, we consider the registration of two 3D point clouds by finding correspondences of four pre-selected points in each scene as illustrated in Fig.~\ref{fig:pointCloudMatching}. 
While this pre-selection is, of course, a very challenging part of the overall solution, our main goal is to illustrate the solution of a (low dimensional) matching problem via quantum computing -- possibly as the solution to a subproblem in an iterative registration algorithm. 

To set up the matching problem, we select four arbitrary points in one scene and use the ground truth transformation between the two frames to identify the corresponding points in the other scene. 
We then set up a matching problem with costs as in \eqref{eq:isometricMatching} by simply using Euclidean distances between the points.

\begin{figure}[htb]
    \centering
    \includegraphics[width = 0.48\linewidth]{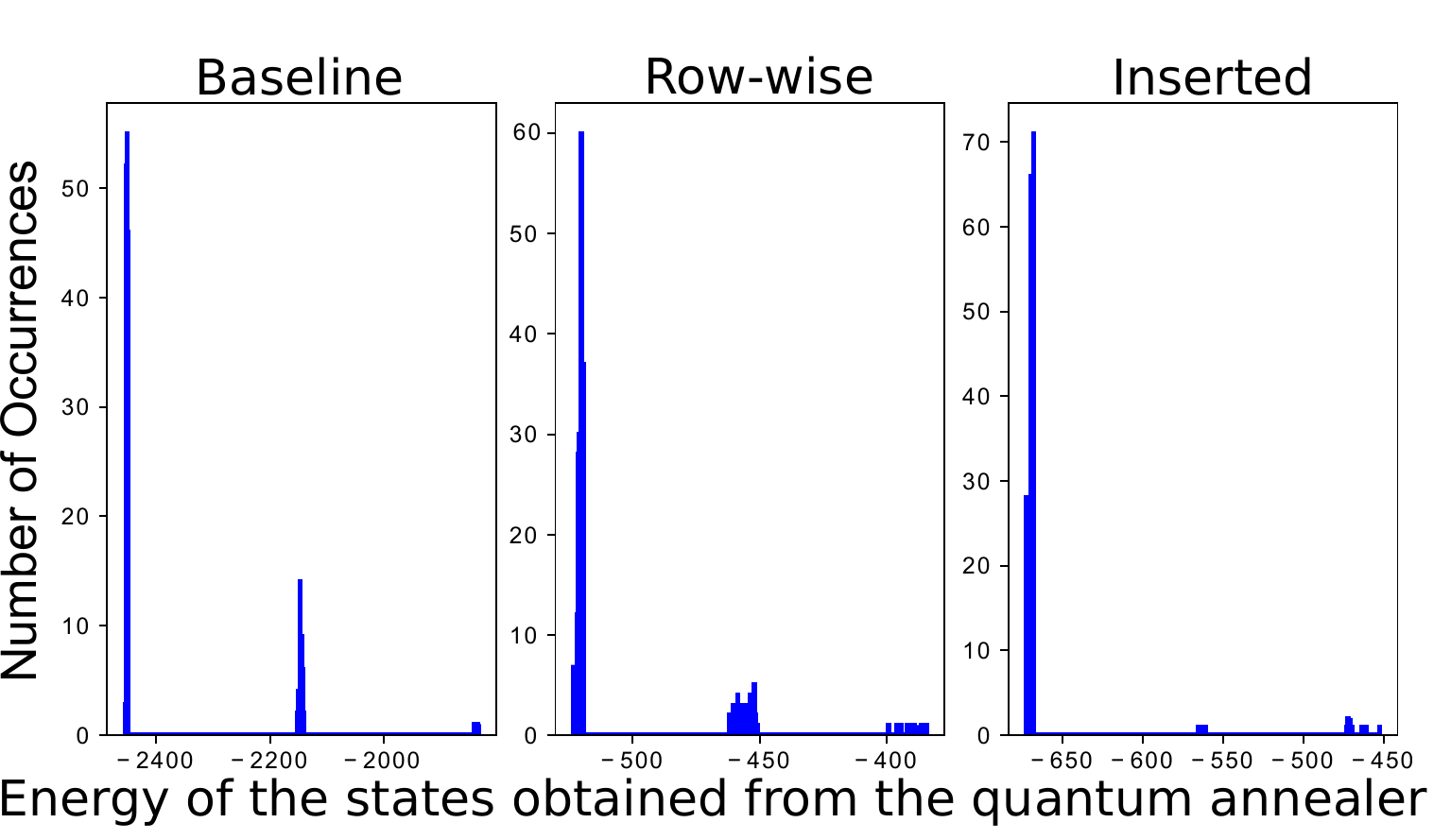}\hspace{10pt}
        \includegraphics[width = 0.48\linewidth]{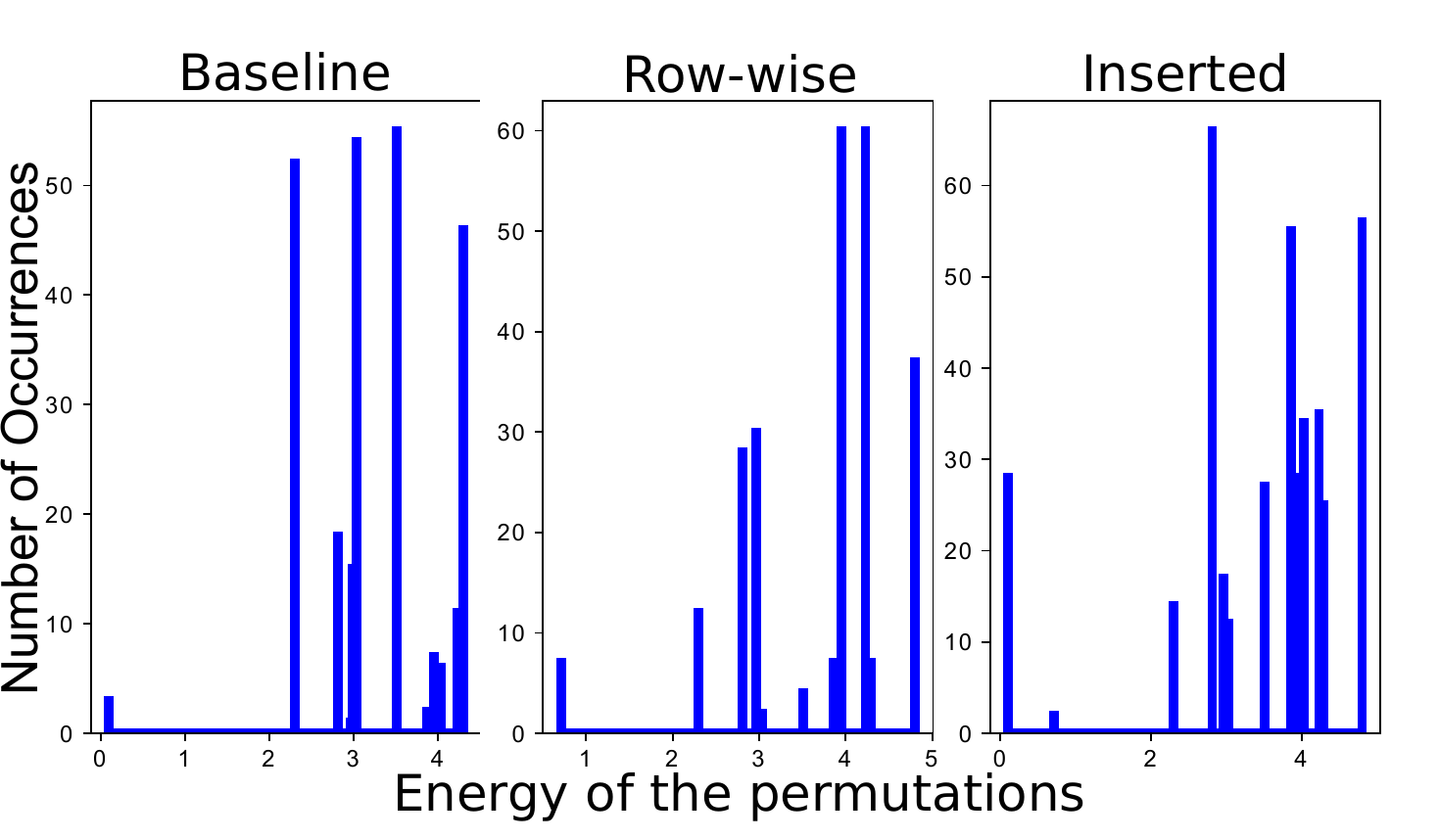}
    \caption{The histograms show the states obtained for the point matching problem in Fig.~\ref{fig:pointCloudMatching}. The left triple shows the overall histogram obtained over 500 anneals while the right triple is a (rescaled) zoom into the leftmost peaks of the left triple, which corresponds to actual permutation matrices. }
    \label{fig:pointCloudMatchingHist}
\end{figure}

The histogram of energies obtained by all three quantum graph matching formulations over $500$ anneals is shown in Fig.~\ref{fig:pointCloudMatchingHist}. 
The top row illustrates the overall histogram with strong peaks at low energies. 
As these peaks correspond to permutation matrices, we can conclude that all penalty terms were successful in strongly promoting permutations, with the inserted formulation yielding the best results. 
Zooming into the leftmost peak (illustrated in the bottom row of Fig.~\ref{fig:pointCloudMatchingHist}), however, reveals that none of the three formulations was successful in consistently predicting the global optimum among the permutation matrices. 
Considering the probabilities of less than $1\%$ for the row-wise and baseline, and about $5.5\%$ for the inserted formulations to predict the ground truth solution, one must conclude that all algorithms do not provide significantly better solutions than random guessing (which has a success probability of $4.17\%$ for $n=4$). 

In summary, this experiment underlines the great difficulty current quantum hardware still has with problem instances of $n=4$, \textit{i.e.,} looking for the values of the best $16$ qubits that are constrained to representing a permutation matrix. 
The performance of our three QUBO formulations on this problem can be seen in Fig.~\ref{fig:pointCloudMatchingHist}. 

\section{Near-Isometric Matching}\label{sec:near_isometric_matching} 
\begin{figure}[htb]
    \centering
    \includegraphics[width = 0.59\textwidth]{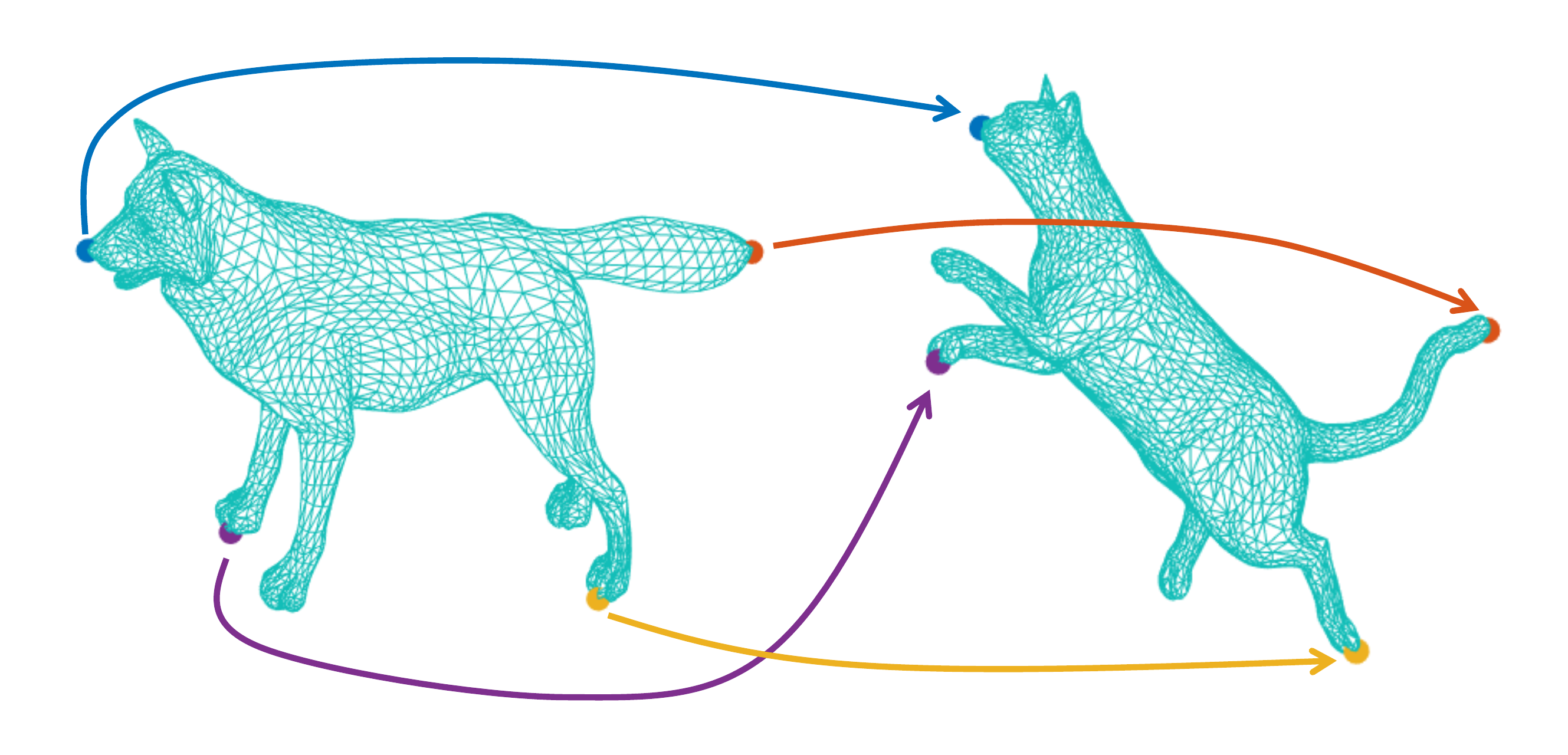}
    \caption{Illustrating a matching problem in which the shapes to be matched are only approximately related by an isometry. }
    \label{fig:nearIsometric}
\end{figure}

Fig.~\ref{fig:nearIsometric} shows an example of a matching problem between two shapes that are only approximately related by an isometry, \textit{i.e.,} matching a wolf to a cat in this particular instance. 
Still modelling the problem as an isometric deformation using costs determined by \eqref{eq:isometricMatching} yields an instance, where the true matching still is the global optimum of the quadratic assignment problem, but where wrong permutations have considerably more similar energies than in the point cloud and isometric shape matching examples. 

\begin{figure}[htb]
    \centering
    \includegraphics[width = 0.51\linewidth]{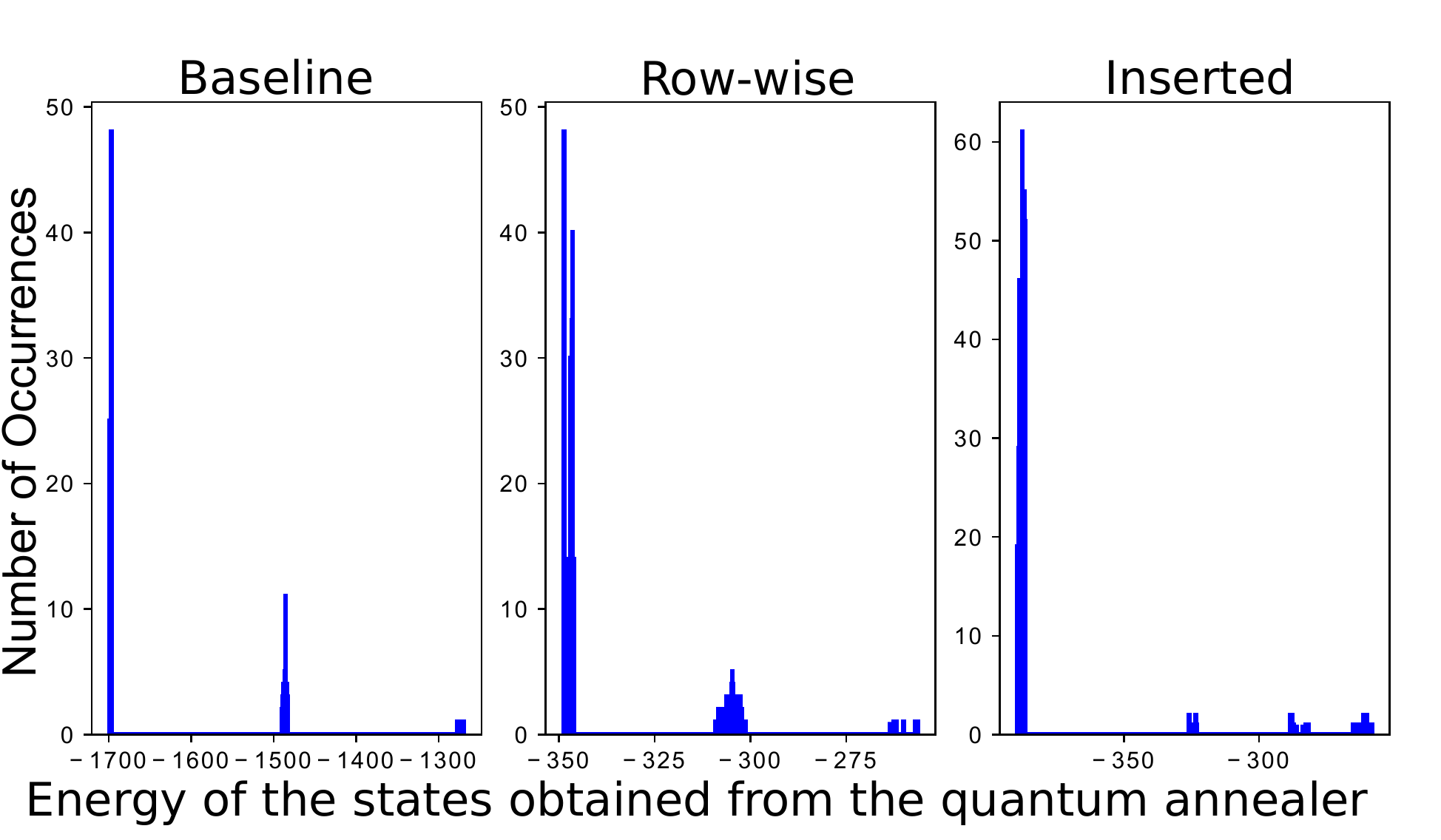}\hspace{10pt}
       \includegraphics[width = 0.44\linewidth]{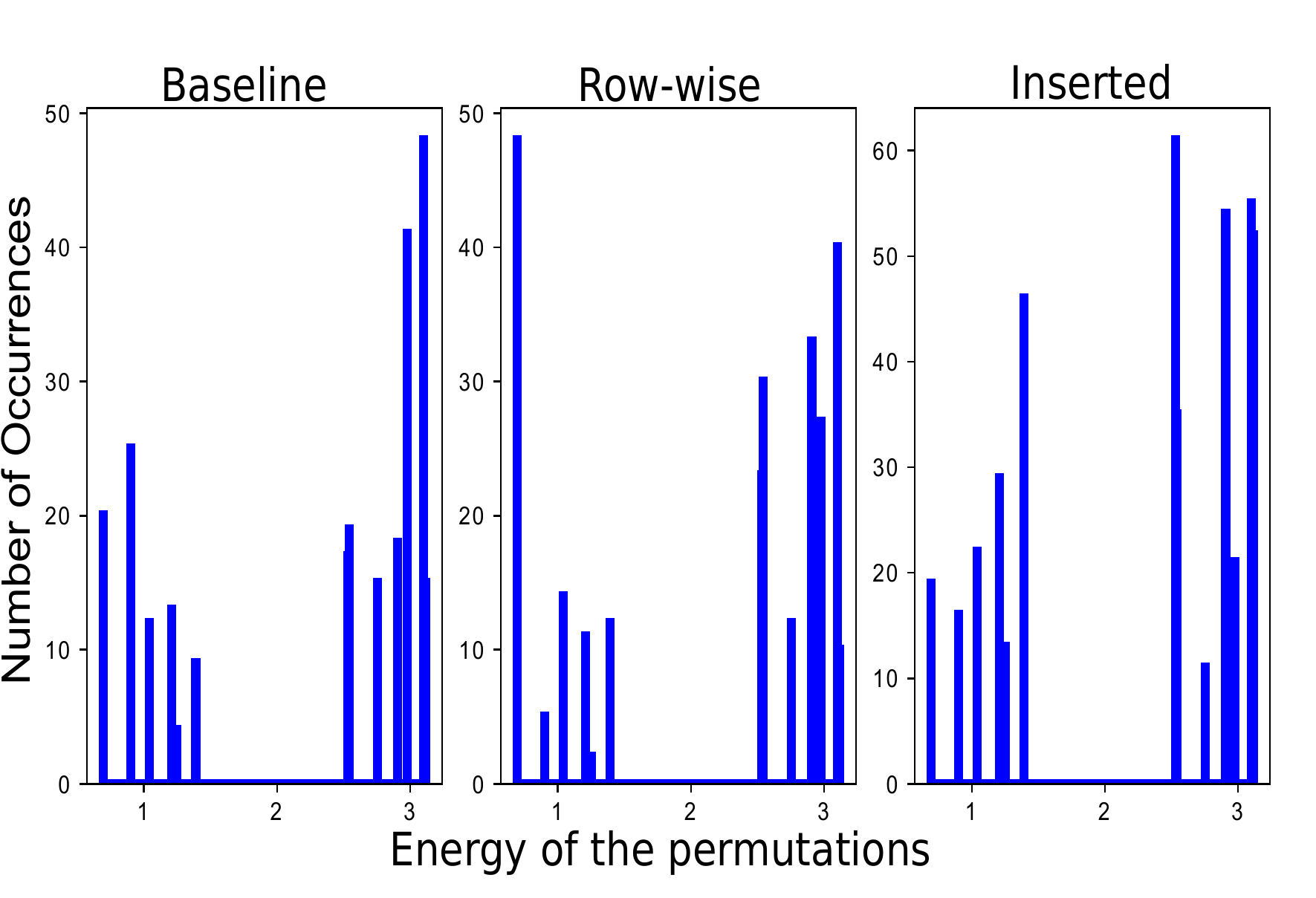}
    \caption{The histograms show the states obtained for the point matching problem in Fig.~\ref{fig:nearIsometric}. The left triple shows the overall histogram obtained over 500 anneals while the right triple is a (rescaled) zoom into the leftmost peaks of the left triple, which corresponds to actual permutation matrices. }
    \label{fig:pointCloudMatchingHistNearIso}
\end{figure}

The success probabilities for the \textit{inserted}, the \textit{baseline} and the \textit{row-wise} QGM variants are $3.8\%$, $4\%$ and $9.2\%$, respectively. 
As we can see in Fig.~\ref{fig:pointCloudMatchingHistNearIso}, the energy landscape has a wide range and the energy values corresponding to permutations are closer to each other compared to the case when the isometry assumption is strictly fulfilled. 
If one looks at the histograms with only permutations, only the \textit{row-wise} QGM shows some trend towards the permutation with the lowest energy. Although its success probability is more than twice as good as random guessing, such a factor to random guessing would still not be sufficient to scale such an algorithm to large $n$.  

\section{Embedding to the Chimera Graph}\label{sec:minor_embeddings} 
\begin{figure*}%
    \centering 
    \includegraphics[width=1\linewidth]{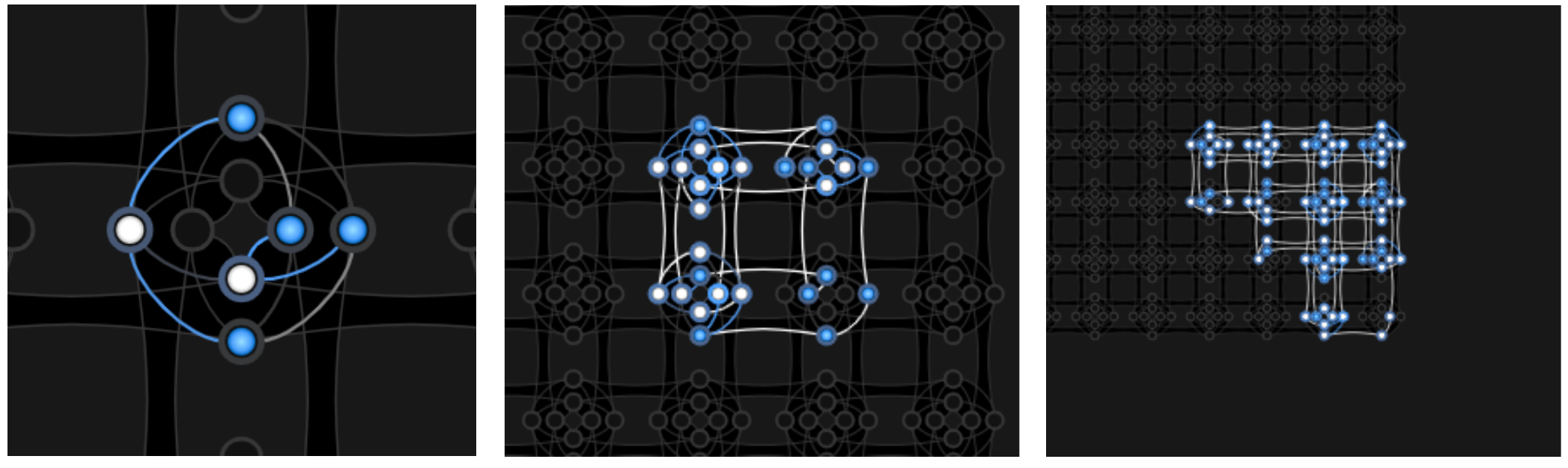} 
    \caption{QGM minor embeddings to the chimera graph, for $n=2$ (left), $n=3$ (middle) and $n = 4$ (right).
    The number of logical qubits grows quadratically with $n$, and the number of physical qubits required to embed the logical qubits to the Chimera graph grows as $n^4$ on the current generation of D-Wave annealers. 
    The white and blue circles denote measured values zero and one, respectively, and the lines between the qubits denote couplers. 
    Grey lines connect the physical qubits with the same resulting values. 
    A subset of those build chains and represent a single logical qubit. 
    Blue lines connect different measured values. (Graphic generated using D-Wave leap problem inspector)
    } 
    \label{fig:embedding} 
\end{figure*}

The embeddings to the Chimera graph for different dimensions with the row-wise formulation can be seen in Fig.~\ref{fig:embedding}. { 
While the number of logical qubits grows quadratically with $n$, the number of physical qubits required to embed those to the Chimera graph grows as $n^4$. 
For $n = 2$, the length of the longest chain is two physical qubits. 
For $n = 3$ and $n = 4$, the chain length does not exceed four and six, respectively. 
}

\section{Discussion of Negative  Results}\label{sec:discussion_negative_results} 
{Since physical quantum computing is an emerging technology, reporting and discussing negative results on the early stage is of high relevance for the community. 
Insights of this section can help in choosing a promising direction for  improvements and future research. 
} 

As discussed in Sec.~\ref{ssec:range_couplings} we claim that the failure to provide the ground state with more probability than random guessing is due to the experimental errors in the coupling parameters. 
To back this up we provide the smallest and largest values of couplings and biases of the regularization and data term in Table \ref{tab:ranges}.

\begin{table}[htbp]
\small 
\centering
\caption{Illustrating the range of values arising from the penalty to constrain each formulation to permutation matrices ($Q_{\text{reg *}}$ and $q_{\text{reg *}}$) and from the actual problem costs ($Q_{\text{prob *}}$ and $q_{\text{prob *}}$). As we can see the constraints contribute more to the quadratic coupling matrix by a factor of around $6$ for the inserted, $13$ for the row-wise, and almost $55$ for the baseline.}
\begin{tabular}{|l|l|l|l|}
\hline
 & Row-wise &Inserted & Baseline \\ \hline
$Q_{\text {max}}$ & -0.887 &-0.987 & -0.973 \\ \hline
$Q_{\text {min}}$ & -1.037&-1.295 & -1.009 \\ \hline
$q_{\text {max}}$ & -1.937&-8.884 & -116.772 \\ \hline
$q_{\text {min}}$ & -2.277&-9.753 & -120.772 \\ \hline
$Q_{\text{reg max}}$ & 0.962&1.121 & 0.983 \\ \hline
$Q_{\text {reg min}}$ & -0.962&-0.705 & -0.991 \\ \hline
$q_{\text {reg max}}$ & 4.017&10.885 & 118.772 \\ \hline
$q_{\text {reg min}}$ & 0.140&7.753 & 118.772 \\ \hline
\end{tabular}
\label{tab:ranges}
\end{table}

The couplings and the biases are scaled, so that they fit the feasible region of the annealer. $Q_{\text{reg}}, q_{\text{reg}}$ do yield constant energies for all permutations and therefore  $Q_{\text{prob}}, q_{\text{prob}}$ contains the information, which permutation is optimal. The sum are the real, physical couplings 
$$Q = Q_{\text{reg}}+Q_{\text{prob}}$$ and biases
$$q= q_{\text{reg}}+q_{\text{prob}}.$$
One can see that most of the accessible range of the coupling parameters has to make sure that the output is a permutation.

There exists already ways to deal with these problems, which we want to try out in further experiments. One possibility would be to use extended $J$-range parameter \cite{DWaveWeitereMoeglichkeiten}. 
The easiest way for this would be to use the virtual graph embedding instead of the default one (EmbeddingComposite). First attempts in this direction show that using the virtual graph embedding requires  a lot of computation time. For one instance, where we got an error warning we had to invest $12\%$ of our access time. 

As reported in the main paper we first used a long annealing path with a break and at some point switched to using 20$\mu s$. Although a longer annealing time can often be used to enhance the success probability, if we additionally look a the time it takes to perform the experiment until one gets the optimum with for example $99 \%$ certainty then according to \cite{king2014algorithm} $20\mu s$ seems to be the better choice, if less than $512$ qubits are used.

\section{Beyond Quantum Computing}
\begin{figure}
\begin{center}
  \hspace{-30pt}\includegraphics[width=0.45\linewidth]{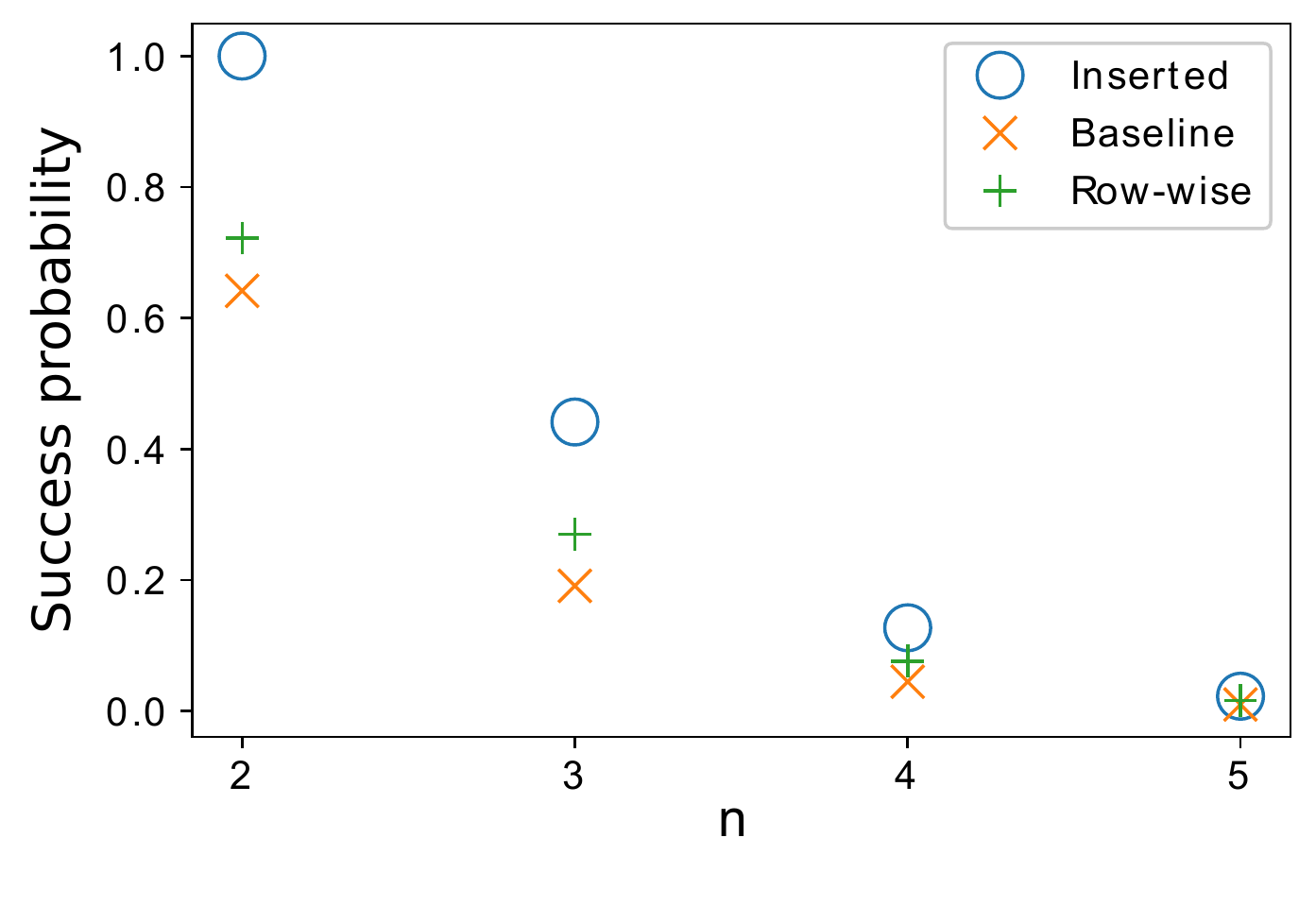}
\end{center}
   \caption{Percentage of the $5000$ runs that found the optimum with simulated annealing averaged over $10$ random problem instances.}
\label{fig:SimulatedAnnealing}
\end{figure}

In addition to the numerical experiments using quantum computing, we also briefly tested the effect of our three reformulations on methods that are inspired by physical systems. In Fig.~\ref{fig:SimulatedAnnealing}, we compare the success probability of our three formulations for simulated annealing using random instances of graph matching problems. As we can see, the row-wise and---even more so---the inserted formulations yield results clearly superior to the baseline method, indicating that our analysis might be of use beyond quantum computing. Due to the $\mathcal{NP}$-hard nature of the underlying problem, it is to be expected that the overall success probability still decreases exponentially with increasing $n$ (for a fixed number of simulated annealing runs).

%% file: theory.tex
In order to ensure that we can incorporate a constraint of the form $A\mathbf{x}=\mathbf{b}$ as a penalty, the corresponding penalty parameter $\lambda$ has to be chosen such that 
\begin{equation} 
\small 
\min_{\mathbf{x}\in \{0,1\}^n, A\mathbf{x} = \mathbf{b}} \mathbf{x}^TW\mathbf{x} < \min_{\mathbf{x}\in \{0,1\}^n, A\mathbf{x} \neq \mathbf{b}} \mathbf{x}^TW\mathbf{x}  + \lambda\|A\mathbf{x}-\mathbf{b}\|^2.
\end{equation} 

A  naive estimate for a sufficiently large $\lambda$ is obtained by using
\begin{align}
     \min_{\mathbf{x}\in \{0,1\}^n, A\mathbf{x} = \mathbf{b}} \mathbf{x}^TW\mathbf{x}  \leq \sum_{i,j}   \max \{0, W_{i,j} \}
\end{align}
and
\begin{align} 
\begin{split}
      &\min_{\mathbf{x}\in \{0,1\}^n, A\mathbf{x} \neq \mathbf{b}} \mathbf{x}^TW\mathbf{x}  + \lambda\|A\mathbf{x}-\mathbf{b}\|^2 
     \geq~ \min_{\mathbf{x}\in \{0,1\}^n, A\mathbf{x} \neq \mathbf{b}} \mathbf{x}^TW\mathbf{x}  + \lambda \min_{\mathbf{x}\in \{0,1\}^n, A\mathbf{x} \neq \mathbf{b}} \|A\mathbf{x}-\mathbf{b}\|^2\\
     &\geq ~\min_{\mathbf{x}\in \{0,1\}^n} \mathbf{x}^TW\mathbf{x}  + \lambda \underbrace{\min_{s\in \{-1,1\}^n, A\mathbf{x} \neq \mathbf{b}} \|A\mathbf{x}-\mathbf{b}\|^2}_{=:d}
     \geq~ - \sum_{i,j}  \max \{0, -W_{i,j} \}  + \lambda d.
\end{split}
\end{align} 
 Thus a sufficient upper bound for $\lambda$ is given by 
 \begin{align*} 
\lambda > \frac{1}{d}\sum_{i,j}(\max \{0, -W_{i,j} \} +\max \{0, W_{i,j} \} ) =  \frac{1}{d}\sum_{i,j}\abs{W_{i,j}}.
\end{align*} 
The influence of the linear term $\mathbf{c}^{\text{T}} \mathbf{x} $ can be calculated in the same way. Alternatively $\mathbf{c}$ can be added to the diagonal of $W$, since $x_i^2= x_i$ for $x_i \in \{ 0,1\}$. 
The fact that $d=2$ can be proven by the following reasoning. Consider that there is a violation for the rows so that one row does not sum up to one and $\sum_{k \text{ belongs to a row} } (Ax-b)_k^2=1$. Since the sum over all rows of a matrix is the same as the sum over all columns, one constraint for the columns is violated as well. Since the same argument can be made if we switch rows and columns $d$ has to be 2.

\subsection{Proof of Proposition {\large \eqref{thm:Reformulation}}} 

Better estimates can be obtained by exploiting more about the specific structure of the constraint. 
We aim to formulate the optimization problem as an unconstrained optimization problem with row-wise penalty parameters:

\begin{equation}
    \label{eq:OptProbMultipleLambda}
    \min_{\mathbf{x}\in \lbrace 0,1\rbrace^{n^2} } \quad  \mathbf{x}^{\text{T}} W \mathbf{x} + \mathbf{c}^{T} \mathbf{x}+ \sum_{i=1}^{2n} \lambda_i ((A\mathbf{x})_i-b_i)^2.
\end{equation}

One can solve the following optimization problems
\begin{equation}
   D_{\mathcal{J}_j}= \max_{k\in \mathcal{J}_j}  (\sum_i  \abs{ (W_{k,i }+ W_{i,k}) } + \abs{ W_{k,k}} + \abs{ c_k})
\end{equation}
for every row or column $\mathcal{J}_j$. 
\\

The following bound for $\lambda_j$ can be used to obtain an unconstrained optimization problem as in \eqref{eq:OptProbMultipleLambda}:
\begin{align}
    \lambda_j=\max_{k\in \mathcal{J}_j}  (\sum_i  \abs{ (W_{k,i }+ W_{i,k}) } + \abs{ W_{k,k}} + \abs{ c_k})   + \frac{1}{2}( \max_{k}  \sum_i  \abs{ (W_{k,i }+ W_{i,k}) } + \abs{ W_{k,k}} + \abs{ c_k})  
    =    D_{\mathcal{J}_j} + \frac{1}{2}  D_{\{1,...,n^2 \} }. 
\end{align}

\begin{proof}
Let $x$ be an arbitrary element in $ \{0,1 \}^{n^2}$. We want to show that for the above choice of $\lambda_j$: 
\begin{align}
    \exists \mathbf{p} \in \text{vec}(\mathbb{P}_n):
       \mathbf{p}^{\text{T}} W \mathbf{p} + \mathbf{c}^{T} \mathbf{p} 
      \leq \mathbf{x}^{\text{T}} W \mathbf{x} + \mathbf{c}^{T} \mathbf{x} +  \sum_{i=1}^{2n} \lambda_i ((A\mathbf{x})_i-b_i)^2. \label{eq:EsGibtPermutation}
\end{align}

For the matrix $X$, which fulfills $x= \text{vec}(X)$, we can construct sets with the property:

\begin{equation} 
\begin{aligned} 
\small 
(i,j)\in I  \Rightarrow  X_{i,j}=1  \quad \wedge \quad & \forall k  \in \{1,...,n\} \setminus \{ j\}  (i,k) \notin I, \\  \wedge \quad  & \forall k  \in \{1,...,n\} \setminus \{ j\}   (k,j) \notin I. 
\end{aligned}
\end{equation} 

We name one of these sets that has the maximal possible number of elements $I_{\text{max}}$. 
The permutation matrix $P$ with $\mathbf{p}=\text{ vec}(P)$ that we want to construct can be any permutation with ones placed at the positions in $I_{\text{max}}$. We can get from $X$ to $P$ by first erasing all ones that are not in the positions $I_{\text{max}}$ and then adding $ n- |I_{\text{max}}|$ ones.

Consider the set of matrices $(X^{(k)})_ {0 \leq k \leq H}$ with:
\begin{equation} 
\begin{aligned}
&  X^{(0)}= X, \\
&  X^{(H)}= P,  \\
&  ||X^{(k)}- X^{(k-1)}|| = 1. 
\end{aligned}
\end{equation} 
These matrices can be constructed if we start from $P$ and erase successively all ones that are not in $I_{\text{max}}$. 
After that we can insert ones that have an index in common with an element in $I_{\text{max}}$. This set we call $B$.

 Inserting or erasing a one at the j-th column yields maximally to an energy difference of 
\begin{equation}
   D_{\mathcal{C}_j}= \max_{k\in \mathcal{C}_j}  (\sum_i  \abs{ (W_{k,i }+ W_{i,k}) } + \abs{ W_{k,k}} + \abs{ c_k}),
\end{equation}
where $\mathcal{C}_j$ describe the indizes that belong to the j-th collumn and analogously $\mathcal{R}_j$ describe the indizes that belong to the j-th row. We define $f$ as:
\begin{equation}
    f (Y ) =        \mathbf{y}^{\text{T}} W \mathbf{y} + \mathbf{c}^{T} \mathbf{y}, 
\end{equation}
with $\mathbf{y}= \text{vec}(Y)$.
To prove \eqref{eq:EsGibtPermutation}, we use the principle of a telescope sum: 
\begin{equation} 
\small 
\begin{aligned}  
    f(P)-f(X) & = \sum_{k=0}^{H-1} f(X^{(k+1)})-f(X^{(k)}) \leq \sum_{k=0}^{H-1} \abs{ f(X^{(k+1)})-f(X^{(k)})} \\
    & = \sum_{\{k\in \{0,...,H-1 \}| \text{pos }(X^{(k+1)}-X^{(k)})\in B \}} \abs{f(X^{(k+1)})-f(X^{(k)})} \\
    & +\sum_{\{k\in \{0,...,H-1 \}| \text{pos }(X^{(k+1)}-X^{(k)})\notin B \}} \abs{f(X^{(k+1)})-f(X^{(k)})}. 
\end{aligned} 
\end{equation} 
For the second sum we want to use: 
\begin{equation}
    \abs{ f(X^{(k+1)})-f(X^{(k)})}\leq D_{\{1,...,n^2\} }. 
\end{equation}
In the first sum we can make use of the specific column or row and use $ D_{ \mathcal{J}_j}$. 
This yields the following ansatz for $\lambda_{\mathcal{R}_j}$, which is the $\lambda_k$ that belongs to the j-th row:
\begin{equation}
\lambda_{\mathcal{R}_j}= \alpha  D_{\mathcal{R}_j} + \beta D_{\{1,...,n^2 \}}.
\end{equation} 
To obtain the constant $\beta$ we have to calculate the right side of the inequality:
\begin{equation}
\beta \geq  \max_{\{ \mathbf{x}\in \{0,1\}^{n^2}| A\mathbf{x} \neq \mathbf{b}\}} \frac{|n-I_{\text{max}}| }{\sum_j \abs{A\mathbf{x}_j - b_j}}.
\end{equation}
If we have a row, where no element of $ I_{\text{max}}$ is present, there can be other ones placed there that share a column with a position in $I_{\text{max}}$. If there is no one in that column, then $\abs{A\mathbf{x}_j - b_j}\neq 0$ for the corresponding j. If there are ones then $\sum_j \abs{A\mathbf{x}_j - b_j}$ also increases, since the ones are in places, where they share a column with a position in $I_{\text{max}}$. This yields to the inequality:
\begin{equation}
\beta \geq  \max_{\{ \mathbf{x}\in \{0,1\}^{n^2}| A\mathbf{x} \neq \mathbf{b}\}} \frac{|n-I_{\text{max}}| }{2 \cdot |n-I_{\text{max}}| } = \frac{1}{2}.
\end{equation}

To get an estimate for $\alpha$ we now consider a row that contains an element of $ I_{\text{max}}$. In the case, that there is only a single one we do not have to delete other ones. Every additional one increases $\abs{A\mathbf{x}_j - b_j}$ also by one. 
Therefore: 
\begin{equation}
\small 
\alpha \geq   \max_{\{j, \mathbf{x}\in \{0,1\}^{n^2}|\left(A\mathbf{x}\right)_j \neq b_j \quad \mathcal{J}_j \cap \text{vec(}I_{\text{max}}) \neq \varnothing \}} \frac{ |\text{vec(}B \setminus I_{\text{max}} ) \cap \mathcal{J}_j   | }{\abs{\left(A\mathbf{x}\right)_j - b_j }} =1.
\end{equation}

\subsection{Proofs of Lemma {\large \eqref{lm:Permutation}}  and Proposition {\large \eqref{thm:ReformulationPermutation}}}
%
\subsubsection{Proof of Lemma \eqref{lm:Permutation}}
\begin{proof}
We express the set of permutation matrices in terms of the coefficients $x_{i,j}$:

\begin{footnotesize}
\begin{align*}
\mathbb{P}_n &= \lbrace X \in \mathbb{R}^{n\times n}  | \forall i,j \in \lbrace 1,...,n \rbrace \; x_{i,j} \in \lbrace 0,1 \rbrace  \quad  \forall j \in \lbrace 1,...,n \rbrace \sum_{i= 1}^n  x_{i,j} =\sum_{i= 1}^n  x_{j,i}= 1   \rbrace \\
&=\Big\lbrace \left( \begin{array}{rrrrr}
1 -\sum_{i= 2}^n  x_{i,1}  & 1 -\sum_{i= 2}^n  x_{i,2} & \dots & 1 -\sum_{i= 2}^n  x_{i,n} \\
x_{2,1} & x_{2,2} & ... &  x_{2,n} \\
\vdots & \vdots & \ddots & \vdots \\
x_{n,1} & x_{n,2} & ... &  x_{n,n} \\
\end{array}\right)  \in \mathbb{R}^{n\times n}  | \forall i \in \lbrace 2,...,n \rbrace , j \in \lbrace 1,...,n \rbrace\text{ }  x_{i,j} \in \lbrace 0,1 \rbrace,  \quad  \\& \forall j \in \lbrace 1,...,n \rbrace \sum_{i= 1}^n  x_{j,i}= 1  \wedge \sum_{i= 2}^n  x_{i,j} \leq 1   \Big\rbrace \\
&=\Big\lbrace \left( \begin{array}{rrrrr}
1 -\sum_{i= 2}^n  x_{i,1}  & 1 -\sum_{i= 2}^n  x_{i,2} & \dots & 1 -\sum_{i= 2}^n  x_{i,n} \\
1- \sum_{i= 2}^n x_{2,i} & x_{2,2} & ... &  x_{2,n} \\
\vdots & \vdots & \ddots & \vdots \\
1- \sum_{i= 2}^n x_{n,i} & x_{n,2} & ... &  x_{n,n} \\
\end{array}\right)  \in \mathbb{R}^{n\times n}  | \forall i \in \lbrace 2,...,n \rbrace , j \in \lbrace 1,...,n \rbrace\text{ }  x_{i,j} \in \lbrace 0,1 \rbrace,  \quad \\ & \forall j \in \lbrace 1,...,n \rbrace \sum_{i= 1}^n  x_{j,i}= 1 \wedge \sum_{i= 2}^n  x_{i,j} \leq 1   \Big\rbrace\\
&= \Big\lbrace     \resizebox{0.45\textwidth}{!}{$ \left( \begin{array}{rrrrr}
1 - (n-1)+\sum_{i,j= 2}^n  x_{i,j}  & 1 -\sum_{i= 2}^n  x_{i,2} & \dots & 1 -\sum_{i= 2}^n  x_{i,n} \\
1- \sum_{i= 2}^n x_{2,i} & x_{2,2} & ... &  x_{2,n} \\
\vdots & \vdots & \ddots & \vdots \\
1- \sum_{i= 2}^n x_{n,i} & x_{n,2} & ... &  x_{n,n} \\
\end{array}\right) $} \in \mathbb{R}^{n\times n}  | \forall i \in \lbrace 2,...,n \rbrace , j \in \lbrace 2,...,n \rbrace\text{ }  x_{i,j} \in \lbrace 0,1 \rbrace,\quad \\ & \sum_{i,j= 2}^n  x_{i,j}  \in \lbrace n-2, n-1 \rbrace  \text{ }  \forall j \in \lbrace 1,...,n \rbrace, \sum_{i= 2}^n  x_{j,i} \leq 1 \wedge   \sum_{i= 2}^n  x_{i,j} \leq 1    \Big\rbrace\\
\end{align*}
\begin{align*}
&=\Big\lbrace \left( \begin{array}{rrrrr}
2 - n+\sum_{i,j= 2}^n  x_{i,j}  & 1 -\sum_{i= 2}^n  x_{i,2} & \dots & 1 -\sum_{i= 2}^n  x_{i,n} \\
1- \sum_{i= 2}^n x_{2,i} & x_{2,2} & ... &  x_{2,n} \\
\vdots & \vdots & \ddots & \vdots \\
1- \sum_{i= 2}^n x_{n,i} & x_{n,2} & ... &  x_{n,n} \\
\end{array}\right) \in \mathbb{R}^{n\times n}  | \forall i \in \lbrace 2,...,n \rbrace , j \in \lbrace 2,...,n \rbrace\text{ }  x_{i,j} \in \lbrace 0,1 \rbrace,\quad \\ & \sum_{i,j= 2}^n  x_{i,j} \in \lbrace n-2, n-1 \rbrace  \text{ }  \forall j \in \lbrace 2,...,n \rbrace \sum_{i= 2}^n  x_{j,i} \leq 1 \wedge   \sum_{i= 2}^n  x_{i,j} \leq 1   \Big \rbrace\\
&=\Big\lbrace \left( \begin{array}{rrrrr}
2 - n+\sum_{i,j= 2}^n  x_{i,j}  & 1 -\sum_{i= 2}^n  x_{i,2} & \dots & 1 -\sum_{i= 2}^n  x_{i,n} \\
1- \sum_{i= 2}^n x_{2,i} & x_{2,2} & ... &  x_{2,n} \\
\vdots & \vdots & \ddots & \vdots \\
1- \sum_{i= 2}^n x_{n,i} & x_{n,2} & ... &  x_{n,n} \\
\end{array}\right)  \in \mathbb{R}^{n\times n}  | \forall i \in \lbrace 2,...,n \rbrace , j \in \lbrace 2,...,n \rbrace\text{ }  x_{i,j} \in \lbrace 0,1 \rbrace,\quad \\ & \sum_{i,j= 2}^n  x_{i,j} \in \lbrace n-2, n-1 \rbrace  \text{ }  \forall j,i,k \in \lbrace 2,...,n \rbrace \quad i\neq k, \quad x_{j,i} x_{j,k} =0 \wedge x_{i,j} x_{k,j} = 0  \Big\rbrace. 
\end{align*}
\end{footnotesize}

\end{proof}

\subsubsection{Proof of Proposition \eqref{thm:ReformulationPermutation}}

We want to find sufficient lower bounds for $\lambda_1^j$ and $\lambda_2$ in

\begin{equation} 
\small 
\begin{aligned}
\min_{\mathbf{x}\in \lbrace 0 ,1\rbrace^{(n-1)^2} } \quad  \mathbf{x}^{\text{T}} \tilde{W} \mathbf{x} + \tilde{c}^{T} \mathbf{x} +  \sum_{j=1}^{2(n-1)} \lambda_1^j \left( \sum_{k \in \mathcal{J}_{j}} x_k  \right) \left( \sum_{k \in \mathcal{J}_{j}} x_k -1\right) 
+ \lambda_2 \left( \sum_{i=1 }^{(n-1)^2}x_i -(n-1)  \right) \left( \sum_{i=1 }^{(n-1)^2}x_i -(n-2) \right) ,
\end{aligned}
\end{equation}
so that it coincides with the constrained optimization problem. 
We make the following ansatz for the $\lambda_2$ parameter:

\begin{equation}
    \lambda_2=  \frac{1}{2} D_{\{1,...,(n-1)^2\}},
\end{equation}
since the function 
\begin{align} 
\left( \sum_{i=1 }^{(n-1)^2}x_i -(n-1)  \right) \left( \sum_{i=1 }^{(n-1)^2}x_i -(n-2) \right) 
\end{align} 
increases faster than $2\cdot H$,when $H$ is the number of entries that need to be switched in order to have $n-1$- or $n-2$-many entries equal to $1$.

This choice for $\lambda_2$ allows us to only investigate the case, where we have $n-1$ or $n$ ones placed anywhere. Similar to the prior proof we ask, what the worst ratio between ones we have to insert anywhere and ones we have to insert or delete in a particular column is.
It can be easily seen that in the worst case scenario all ones are in one column or row.
Therefore the ratio is one half. This shows that we can choose: 

\begin{equation}
    \lambda_1^{j}=  \frac{1}{2} D_{\mathcal{J}_{j}} + \frac{1}{2}D_{\{1,...,(n-1)^2\}}.
\end{equation}

\end{proof}